\documentclass{article}

\usepackage{microtype}
\usepackage{graphicx}
\usepackage{subfigure}
\usepackage{booktabs} %
\usepackage{multirow}
\usepackage{threeparttable}
\usepackage{colortbl}
\usepackage[dvipsnames]{xcolor}
\usepackage[most]{tcolorbox}
\usepackage[hang,flushmargin]{footmisc}
\usepackage{fontawesome}

\usepackage{hyperref}

\usepackage[accepted]{icml2025}

\usepackage{amsmath}
\usepackage{amssymb}
\usepackage{mathtools}
\usepackage{amsthm}
\usepackage{inconsolata}

\usepackage[capitalize,noabbrev]{cleveref}
\usepackage{xspace}

\theoremstyle{plain}
\newtheorem{theorem}{Theorem}[section]
\newtheorem{proposition}[theorem]{Proposition}

\theoremstyle{definition}

\theoremstyle{remark}

\newcommand{\captionvskip}{0.1in}

\newcommand{\smalltt}[1]{{\small\texttt{#1}}\xspace}
\newcommand{\wiki}{\textsc{Wiki}\xspace}
\newcommand{\bills}{\textsc{Bills}\xspace}
\newcommand{\headlines}{\textsc{Headlines}\xspace}
\newcommand{\yelp}{\textsc{Yelp}\xspace}
\newcommand{\congress}{\textsc{Congress}\xspace}
\newcommand{\embeddingmodel}{text-embedding-3-small\xspace}
\newcommand\hyp[1]{\textit{#1}}

\newcommand{\ourmethod}{\textsc{HypotheSAEs}\xspace}
\newcommand{\nlparam}{\textsc{NLParam}\xspace}
\newcommand{\hypogenic}{\textsc{HypoGeniC}\xspace}
\newcommand{\bertopic}{\textsc{BERTopic}\xspace}

\newcommand{\RR}{\mathbb{R}}
\newcommand{\EE}{\mathbb{E}}

\DeclareMathOperator{\topk}{TopK}
\DeclareMathOperator{\auxk}{AuxK}
\DeclareMathOperator{\enc}{enc}
\DeclareMathOperator{\dec}{dec}
\DeclareMathOperator{\relu}{ReLU}

\usepackage{bm}

\newcommand{\papertitle}{Sparse Autoencoders for Hypothesis Generation}

\usepackage[textsize=tiny]{todonotes}
\usepackage{enumitem}
\setlist[itemize]{leftmargin=*}

\icmltitlerunning{\papertitle}

\begin{document}

\twocolumn[
\icmltitle{\papertitle}

\icmlsetsymbol{equal}{*}

\begin{icmlauthorlist}
\icmlauthor{Rajiv Movva}{equal,berkeley}
\icmlauthor{Kenny Peng}{equal,cornell,tech}
\icmlauthor{Nikhil Garg}{tech}
\icmlauthor{Jon Kleinberg}{cornell}
\icmlauthor{Emma Pierson}{berkeley}
\end{icmlauthorlist}

\icmlaffiliation{berkeley}{UC Berkeley}
\icmlaffiliation{cornell}{Cornell University}
\icmlaffiliation{tech}{Cornell Tech}

\icmlcorrespondingauthor{Rajiv Movva}{rmovva@berkeley.edu}
\icmlcorrespondingauthor{Kenny Peng}{kennypeng@cs.cornell.edu}

\icmlkeywords{Machine Learning, ICML}

\vskip 0.3in
]

\printAffiliationsAndNotice{}  %

\begin{abstract}
    We describe \ourmethod, a general method to hypothesize interpretable relationships between text data (e.g., headlines) and a target variable (e.g., clicks). \ourmethod has three steps: (1) train a sparse autoencoder on text embeddings to produce interpretable features describing the data distribution, (2) select features that predict the target variable, and (3) generate a natural language interpretation of each feature (e.g., \textit{mentions being surprised or shocked}) using an LLM. Each interpretation serves as a hypothesis about what predicts the target variable.
    Compared to baselines, our method better identifies reference hypotheses on synthetic datasets (at least +0.06 in F1) and produces more predictive hypotheses on real datasets ($\sim $twice as many significant findings), despite requiring 1-2 orders of magnitude less compute than recent LLM-based methods. 
    \ourmethod also produces novel discoveries on two well-studied tasks: explaining partisan differences in Congressional speeches and identifying drivers of engagement with online headlines.
\end{abstract}

\section{Introduction}
\label{sec:intro}

Large language models (LLMs) show promise as a tool for \textit{hypothesis generation}. Discovering relationships between text data and a target variable is an important and fundamental task with diverse applications in economics, political science, sociology, medicine, and business \citep{grimmer2010bayesian, rathje2021out, sun_negative_2022, gentzkow_what_2010, monroe_fightin_2009, nelson2020computational, ranard2016yelp, ting2017using}. What features of a restaurant review predict a low rating? What features of a social media post predict whether it will go viral? What features of a patient's clinical notes predict if they will develop cancer? 

Automated approaches to answering these questions have the potential to significantly expand and accelerate scientific discovery \citep{ludwig_machine_2023a}. Indeed, multiple lines of work---spanning decades of research---have sought to extract text features that can predict a target variable \citep{blei_latent_2003, monroe_fightin_2009}. Recent LLM-based hypothesis generation methods are especially exciting since they operate directly at the level of human-interpretable natural language concepts \citep{zhou_hypothesis_2024, batista_words_2024}.
Concretely, hypothesis generation methods take a dataset of texts linked to a target variable (e.g., headlines and their engagement level) and hypothesize natural language concepts that predict the target variable (e.g., \textit{mentions being surprised or shocked} or \textit{asks a question to the reader}).

However, basic hurdles impede the full usage of LLMs for hypothesis generation. Consider two natural approaches. The first prompts an LLM with positive and negative examples and asks it to hypothesize concepts that distinguish them \citep{zhou_hypothesis_2024, batista_words_2024, ludan_interpretabledesign_2024}. However, due to context window and reasoning limitations, we can only give LLMs a few training examples at a time, making this hard to scale. A second approach aims to interpret a model fine-tuned to predict the target variable \citep{zhong_explaining_2024}, leveraging the full training dataset. However, neurons are often hard to interpret \citep{elhage_toy_2022}. In summary, efficient and scalable learning of statistical relationships requires numerical representations of text (e.g., \textit{neurons}), but interpreting neurons is hard.

\begin{figure*}
\begin{center}
    \includegraphics[width=0.84\linewidth]{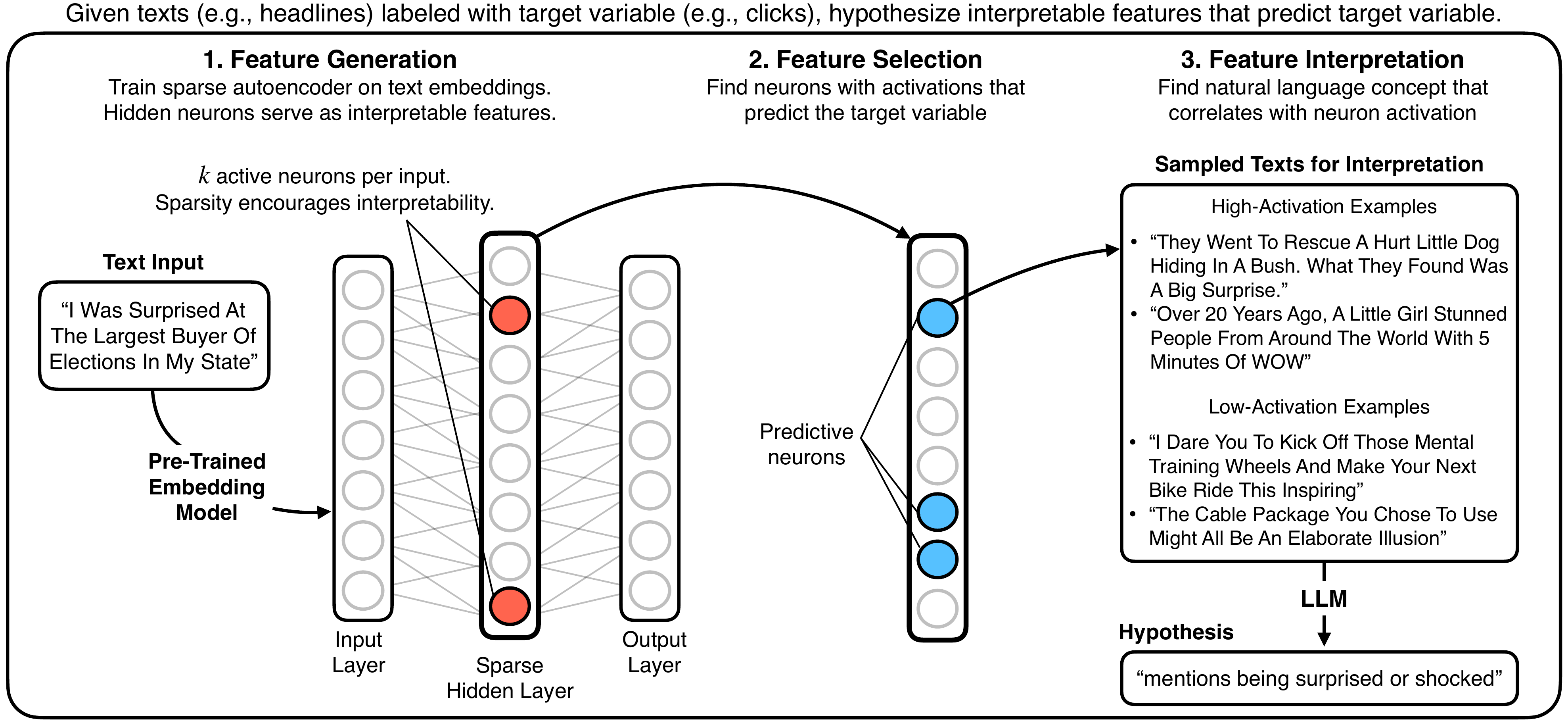}
\end{center}
    \caption{\ourmethod is a general method that outputs natural language hypotheses from text datasets.}
\end{figure*}

Our first contribution is a theoretical framework that connects model interpretation and hypothesis generation, clarifying the challenge above. \Cref{prop:errors-in-variables} gives a ``triangle inequality'' for hypothesis generation: if a neuron is predictive of the target variable, then a sufficiently high-fidelity interpretation of the neuron is also predictive. This result directly motivates a general hypothesis generation procedure: 
\begin{enumerate}
    \item (Feature generation) Learn interpretable neurons (i.e., that tend to fire in the presence of human concepts).
    \item (Feature selection) Select neurons that predict the target variable.
    \item (Feature interpretation) Generate high-fidelity natural language interpretations of these neurons.   
\end{enumerate}

Our second contribution is a method, \ourmethod, that successfully implements this procedure. A primary challenge is learning the interpretable neurons in step 1---after all, a long line of work has established the challenge of interpreting neural networks \citep{olah_feature_2017, kim_interpretability_2018, elhage_toy_2022}. To overcome this, we train a sparse autoencoder (SAE; \citet{makhzani_ksparse_2014}) on pre-trained text embeddings from our dataset. Neurons in the SAE hidden layer are highly interpretable, while retaining the explanatory power of embeddings \citep{cunningham2023sparse, bricken2023towards, templeton2024scaling, gao_scaling_2024, oneill_disentangling_2024}. Intuitively, this architecture leverages the fact that natural language is sparse; only a small fraction of all human concepts are expressed at a time. In step 2, we select neurons that are predictive of the target variable (e.g., using Lasso). In step 3, we automatically generate high-fidelity interpretations of these predictive neurons by prompting an LLM with texts that activate the neuron, building on recent work in autointerpretability \citep{bills2023language}. These interpretations serve as hypotheses.

We evaluate \ourmethod against recent state-of-the-art methods. We consider three synthetic tasks, as well as three real-world tasks of practical interest: hypothesizing the relationship between headlines and engagement, speech text and political party, and review text and rating. Example \ourmethod hypotheses are given below:

\begin{center}
\small
\begin{tabular}{p{1.7cm}p{5.7cm}}
\toprule
Task & Example Hypotheses (Abbreviated) \\ \midrule
\multirow{2}{2cm}{Headline Engagement}          & (+) \hyp{mentions being surprised or shocked} \\
 & (-) \hyp{asks a question directly to the reader} \\ 
\midrule
\multirow{2}{2cm}{Speaker Party}          & (Rep.) \hyp{discusses illegal immigration} \\
 & (Dem.) \hyp{criticizes tax breaks for the wealthy} \\ 
\midrule
\multirow{2}{2cm}{Restaurant Rating}          & (+) \hyp{mentions plans to return to the restaurant} \\
 & (-) \hyp{mentions issues related to food safety} \\ 
\bottomrule
\end{tabular}
\end{center}

\ourmethod produces many more significant hypotheses than three baseline methods. On three real-world tasks, 45/60 hypotheses generated by our method are significant, compared to at most 24 for the baselines. On two datasets, we identify new hypotheses which previous detailed analyses have not uncovered. Intuitively, by decoupling feature generation from the prediction task, \ourmethod is able to explore a wider range of hypotheses.

\ourmethod is also more than 10$\times$ faster and cheaper than recent LLM baselines, with similar runtime and cost to \bertopic, a standard embedding-based topic model. Filtering hypotheses based on the predictiveness of neurons is significantly cheaper than directly evaluating natural language concepts, since all neuron activations of an input can be computed from a single forward pass of an SAE (in comparison to one LLM call per natural language feature).

\section{Related Work}
\label{sec:relatedwork}

\subsection{Hypothesis Generation}

Classical text analysis methods, such as comparing $n$-gram frequencies between groups (Fightin' Words; \citet{monroe_fightin_2009}) or topic modeling (LDA, \citet{blei_latent_2003}; BERTopic, \citet{grootendorst_bertopic_2022}) remain standard tools to discover relationships between text and target variables \citep{grimmer_text_2022}.
A challenge with these methods is that their outputs---lists of words or documents---are not immediately interpretable by humans \citep{chang_reading_2009}.
As a result, applying these methods for scientific discovery requires sufficient domain expertise to prespecify hypotheses and assess whether the data support them \citep{demszky_analyzing_2019, gentzkow_measuring_2016, sun_negative_2022}.

Recently, several LLM-based methods aim to automate interpretable concept discovery from text datasets. \citet{grootendorst_llm_2024} prompts LLMs to generate cluster labels, \citet{pham_topicgpt_2024} use LLMs for end-to-end topic modeling in natural language, and \citet{lam2024conceptInduction} build an LLM-driven interface to discover topics guided by user input.
Another literature uses LLMs to propose concepts in bottleneck models \citep{koh_concept_2020}:
\citet{ludan_interpretabledesign_2024} prompt LLMs with a task description to generate predictive concepts, and \citet{feng_bayesian_2024} use LLMs to provide a prior distribution over concepts. 
\citet{sun_concept_2024} and \citet{zhong_explaining_2024} also use LLMs to propose concepts, guided by neurons from a supervised model.
Finally, \citet{zhou_hypothesis_2024} and \citet{batista_words_2024} focus specifically on hypothesis generation: LLMs propose natural language descriptions of predictive relationships, which can then be tested for generalization.

Unlike several of these methods, \ourmethod decouples feature generation (unsupervised, using the SAE), selection (supervised), and interpretation (the only LLM-dependent step). 
Comparing to baselines from each class of methods (\S\ref{sec:baselines}), we illustrate quantitative and qualitative improvements over classical and fully-LLM-driven baselines.

\subsection{Sparse Autoencoders}

A recent line of work demonstrates that sparse autoencoders \citep{makhzani_ksparse_2014} are an effective architecture to learn interpretable features from intermediate layers of large language models \citep{cunningham2023sparse, bricken2023towards, templeton2024scaling, bills2023language}. This literature is primarily motivated by mechanistic interpretability: since the neurons in the hidden layer of sparse autoencoders correspond to interpretable features, they can be strategically altered to steer the behavior of language models. In the present work, we instead apply sparse autoencoders to learn interpretable text features for the purpose of hypothesizing interpretable relationships between text data and a target variable. We follow \citet{oneill_disentangling_2024}, who show that SAE features can guide semantic search applications, in training a sparse autoencoder directly on text embeddings.

\section{Theoretical Framework}
\label{sec:theory}

In this section, we derive a theoretical result that shows how a language model with highly interpretable neurons can be used for hypothesis generation, as well as where performance loss arises in such a procedure.

Let $Z$ be an indicator for whether a neuron fires for a given text input (i.e., whether the activation exceeds a threshold). Let $\hat{Z}$ be an indicator for whether the text contains a specified natural language concept. Let $Y$ be the target variable. Define the predictiveness of $\hat{Z}$ as the separation score $$S(\hat{Z}):=|\EE[Y|\hat{Z}=1] - \EE[Y|\hat{Z}=0]|.$$ Define $S(Z)$ analogously.\footnote{Separation score is one way we evaluate hypotheses empirically. We suspect that analogs of \Cref{prop:errors-in-variables} hold for alternative measures of predictiveness, like explained variance.} 
We measure the fidelity of the interpretation $\hat{Z}$ by 
\begin{equation}
    \delta(\hat{Z}, Z) := \frac{1 - \min\left\{\text{recall}, \text{precision}\right\}}{\min\{\Pr[\hat{Z}=0],\Pr[Z=0]\}},
\end{equation}
where precision is how often the neuron fires when the concept is present and recall is how often the concept is present when the neuron fires. Lower $\delta$ implies higher fidelity. High recall and precision means the interpretation approximates the neuron's behavior, and is not overly narrow or generic. Empirically, the denominator $\min\{\Pr[\hat{Z}=0],\Pr[Z=0]\}$ is close to 1 (neurons and concepts activate infrequently), so maximizing fidelity is approximately maximizing the minimum of recall and precision.

\begin{proposition}[A Triangle Inequality for Hypothesis Generation]\label{prop:errors-in-variables}
Suppose $Y$ is supported on $[0,1]$. Then 
\begin{align}
    |S(\hat{Z}) - S(Z)| \le \delta(\hat{Z}, Z).
\end{align}
\end{proposition}
Intuitively, the result shows that if $\hat{Z}$ is a high-fidelity predictor of $Z$, then it will also have a similar separation score. A full proof is given in \Cref{sec:proof}.\footnote{There are two main observations that enable us to show the result. First, we focus on $|\EE[Y|\hat{Z}=1]-\EE[Y|Z=1]|,$ and show that if the false negative rate (FNR) is greater than the false discovery rate (FDR), this value is bounded by the FNR; otherwise, it is bounded by the FDR. Second, we show roughly that as long as $\Pr[\hat{Z}=1], \Pr[Z=1] < \Pr[\hat{Z}=0], \Pr[Z=0],$ then $|\EE[Y|\hat{Z}=0]-\EE[Y|Z=0]| < |\EE[Y|\hat{Z}=1]-\EE[Y|Z=1]|.$}

Notice that what we ultimately care about is $S(\hat{Z})$ (how predictive our actual natural language hypothesis is), but our procedure focuses primarily on generating, selecting, and then interpreting the neuron $Z$. \Cref{prop:errors-in-variables} shows that as long as we can generate a sufficiently high-fidelity interpretation $\hat{Z}$ of each neuron $Z$, then we can focus on finding predictive neurons as a way to ultimately find natural language hypotheses. In particular, the result shows that
\begin{align}
    S(\hat{Z}) &= S(Z) + (S(\hat{Z}) - S(Z))\\
    &\ge \underbrace{S(Z)}_{\scriptsize\shortstack{neuron\\predictiveness}} - \underbrace{\delta(\hat{Z}, Z)}_{\scriptsize\shortstack{interpretation\\fidelity}}.\label{eq:decomposition}
\end{align}
\Cref{eq:decomposition} shows that to find a natural language concept such that $S(\hat{Z})$ is large, it suffices to first identify $Z$ such that $S(Z)$ is high (i.e., find a predictive neuron) and then $\hat{Z}$ such that $\delta(\hat{Z}, Z)$ is small (i.e., find a high-fidelity interpretation).

This directly motivates the actual procedure employed in \ourmethod. First, we learn interpretable neurons $Z$ using an SAE. Second, we identify which neurons $Z$ are predictive. Third, we generate high-fidelity interpretations $\hat{Z}$ of these neurons with an LLM.

\Cref{eq:decomposition} also gives us a lens by which to analyze \ourmethod empirically. The result decomposes the performance of hypotheses $\hat{Z}$ into two parts: the predictiveness of the neurons $Z$ and the fidelity of the interpretations $\hat{Z}.$ For example, if the SAE neurons are similarly predictive to the text embeddings they encode, then most performance loss (compared to predicting directly from embeddings) comes from low-fidelity neuron interpretation. Imperfect neuron interpretations can be due to either neurons being fundamentally uninterpretable (there doesn't \textit{exist} a high-fidelity interpretation), or due to difficulties in actually \textit{finding} the interpretation. These challenges arise in steps 1 (training the SAE) and step 3 (generating the interpretation) respectively. We perform such an empirical analysis in \Cref{sec:performance_losses}.

\section{Methods}
\label{sec:methods_main}

In the previous section, we established theoretically that we can generate hypotheses by learning interpretable neurons, identifying which neurons are predictive, and then generating high-fidelity interpretations of the neurons. We now describe our method, \ourmethod, which accomplishes these steps in practice.

Given a dataset of training examples $\{(x_i, y_i)\}_{i\in [N]},$ where $x_i$ is the input text and $y_i$ is the target variable annotation, \ourmethod outputs $H$ natural language concepts that serve as hypotheses. The goal is for these natural language concepts to predict the target variable.

\subsection{Feature Generation: Training Sparse Autoencoders.}

Let $e_i$ denote a $D$-dimensional text embedding of $x_i$. We train a $k$-sparse autoencoder \citep{makhzani_ksparse_2014}. $k$-sparse autoencoders have successfully generated interpretable features both when applied to intermediate layers of language models, as well as text embeddings \citep{gao_scaling_2024, oneill_disentangling_2024}. The SAE encodes a text embedding $e_i$ as follows (we use OpenAI's \embeddingmodel, where $D=1536$, in our experiments):
\begin{align}
    z_i &= \relu(\topk(W_{\enc}(e_i - b_{\mathrm{pre}}) + b_{\enc})),\\
    \hat{e_i} &= W_{\dec}z_i + b_{\dec},
\end{align}
where $b_{\mathrm{pre}} \in \RR^D, W_{\enc}\in \RR^{M\times D}, b_{\enc}\in \RR^{M}, W_{\dec}\in \RR^{D\times M}, b_{\dec}\in \RR^{D},$ and $\topk$ sets all activations except the top $k$ to zero. In a basic $k$-sparse autoencoder, the loss on one input is
\begin{equation}
\mathcal{L} = ||e_i - \hat{e_i}||_2^2.
\end{equation}
We follow \citet{gao_scaling_2024, oneill_disentangling_2024} in further adding an auxillary loss to avoid ``dead latents.'' We describe details and hyperparameters in Appendix \ref{sec:hyperparams}. 

Empirically, the dimensions of $z_i$ are often highly interpretable, corresponding to human concepts.
$M$ sets the total number of concepts learned across the entire dataset, and $k$ sets the number of concepts that can be used to reconstruct each instance. The output of this first step is an $N\times M$ activation matrix, $Z_{\text{SAE}}$.

\subsection{Feature Selection: Target Prediction with SAE Neurons.}
To select a predictive subset of SAE neurons, we fit an $L_1$-regularized linear or logistic regression predicting the target variable $Y$ from the activation matrix $Z_{\text{SAE}}$ \cite{tibshirani_regression_1996}. 
Formally, for regression\footnote{For classification we analogously use an $L_1$-regularized loss: $\mathcal{L}(\bm{\beta}; \lambda) = \frac{1}{N} \operatorname{BCE}(\mathbf{y}, \sigma(Z_{\text{SAE}}\bm{\beta})) + \lambda ||\bm{\beta}||_1$, where $\operatorname{BCE}$ is the binary cross-entropy loss and $\sigma(\cdot)$ is the sigmoid function.}, we optimize $$\min_{\bm{\beta}} \mathcal{L}(\bm{\beta}; \lambda) = \frac{1}{N} \left||\mathbf{y} - Z_{\text{SAE}}\bm{\beta}|\right|^2_2 + \lambda ||\bm{\beta}||_1,$$ where $\mathbf{y}$ is a length-$N$ vector, $Z_{\text{SAE}}$ is an $N \times M$ matrix, and $\bm{\beta}$ is a length-$M$ vector of feature coefficients for each SAE neuron.
The $L_1$ penalty produces sparse coefficient vectors $\bm{\beta}$ where some coefficients are exactly zero (indicating a dropped feature). 
To generate $H$ hypotheses, we perform binary search to identify a value of $\lambda$ which produces exactly $H$ nonzero coefficients.

\subsection{Feature Interpretation: Labeling Neurons with LLMs.}\label{sec:methods-interpretations}

In this step, we generate high-fidelity interpretations (i.e., natural language concepts) of the subset of predictive neurons. 
To generate interpretations of a neuron, we prompt an LLM with example texts from a range of neuron activations and instruct it to identify a natural language concept that is present in high-activation texts and absent in low-activation texts. We generate multiple interpretations by running this procedure several times, and choose the best interpretation according to a measure of fidelity we describe below.

We test a variety of approaches for generating interpretations inspired by recent literature on automated interpretability \citep{bills2023language, templeton2024scaling}. In our final procedure, we sample high- and low-activating texts from the neuron's positive activation distribution, with the sampling percentile bins given in \Cref{sec:hyperparams_neuronlabeling}. 
(Compared to zero-activating texts, texts with a low but nonzero activation act as hard negatives, improving the generated concept's precision.) 
Treating these as true positives and true negatives, respectively, we prompt an LLM to generate a natural language concept that distinguishes 10 samples of each class. 
We then define an interpretation's fidelity as its F1 score in this prediction task, using an LLM to annotate 100 positive and 100 negative samples for the presence of the concept. 
We use GPT-4o with temperature 0.7 to generate concepts and GPT-4o-mini with temperature 0 for concept annotation.

This approach produced predictive interpretations when compared to other prompting and evaluation schemes, though some other methods performed similarly (see \Cref{sec:autointerp_expts} for details). Importantly, we found experimentally that higher-fidelity interpretations according to our measure have better predictive performance (\Cref{sec:fidelity_predictiveness}), justifying our approach to selecting interpretations.

\section{Experiments}
\label{sec:experiments}

\begin{table*}[htbp]
\centering
\scriptsize
\renewcommand{\arraystretch}{1.2}
\definecolor{darkergreen}{HTML}{c8e6c9}
\definecolor{lightgreen}{HTML}{e8f5e9}
\definecolor{darkerred}{HTML}{ffd1d1}
\definecolor{lightred}{HTML}{ffe3e3}

\begin{tabular}{lccp{8cm}rr}
\toprule
Source & Overall AUC & \# Sig & Significant Hypotheses: \colorbox{darkergreen}{Green: $\uparrow$ clicks}; \colorbox{darkerred}{Red: $\downarrow$ clicks} & Sep. & AUC \\
\midrule
\multirow{15}{*}{\ourmethod}& \multirow{15}{*}{0.693} & \multirow{15}{*}{15} & 
\cellcolor{darkergreen}\hyp{mentions photography, images, or visual representations} & \cellcolor{darkergreen}0.26 & \cellcolor{darkergreen}0.51 \\
& & & \cellcolor{lightgreen}\hyp{mentions situations or behaviors that evoke discomfort or embarrassment} & \cellcolor{lightgreen}0.16 & \cellcolor{lightgreen}0.58 \\
& & & \cellcolor{darkergreen}\hyp{includes a reference to being surprised, shocked, or unprepared} & \cellcolor{darkergreen}0.16 & \cellcolor{darkergreen}0.56 \\
& & & \cellcolor{lightgreen}\hyp{focuses on a female subject and her personal journey} & \cellcolor{lightgreen}0.14 & \cellcolor{lightgreen}0.52 \\
& & & \cellcolor{darkergreen}\hyp{mentions or heavily emphasizes a video or video-related content} & \cellcolor{darkergreen}0.13 & \cellcolor{darkergreen}0.53 \\
& & & \cellcolor{lightgreen}\hyp{mentions a surprising or unexpected action or outcome} & \cellcolor{lightgreen}0.13 & \cellcolor{lightgreen}0.58 \\
& & & \cellcolor{darkergreen}\hyp{focuses on a male protagonist or subject} & \cellcolor{darkergreen}0.11 & \cellcolor{darkergreen}0.53 \\
& & & \cellcolor{lightgreen}\hyp{contains phrases about looking or seeing something} & \cellcolor{lightgreen}0.09 & \cellcolor{lightgreen}0.52 \\
& & & \cellcolor{darkergreen}\hyp{mentions uniqueness or exclusivity using superlative or comparative language} & \cellcolor{darkergreen}0.09 & \cellcolor{darkergreen}0.53 \\
& & & \cellcolor{lightgreen}\hyp{mentions a response to or confrontation with offensive behavior} & \cellcolor{lightgreen}0.09 & \cellcolor{lightgreen}0.53 \\
& & & \cellcolor{darkerred}\hyp{mentions politicians, government actions, or political topics} & \cellcolor{darkerred}-0.10 & \cellcolor{darkerred}0.52 \\
& & & \cellcolor{lightred}\hyp{asks a question directly to the reader} & \cellcolor{lightred}-0.12 & \cellcolor{lightred}0.53 \\
& & & \cellcolor{darkerred}\hyp{addresses collective human responsibility or action} & \cellcolor{darkerred}-0.13 & \cellcolor{darkerred}0.56 \\
& & & \cellcolor{lightred}\hyp{mentions environmental issues or ecological consequences} & \cellcolor{lightred}-0.15 & \cellcolor{lightred}0.51 \\
& & & \cellcolor{darkerred}\hyp{mentions actions or initiatives that positively impact a community} & \cellcolor{darkerred}-0.16 & \cellcolor{darkerred}0.53 \\
\midrule
\multirow{1}{*}{\bertopic} & \multirow{1}{*}{0.619} & \multirow{1}{*}{1} & 
\cellcolor{darkergreen}\hyp{headlines emphasize emotional reactions to short, impactful videos} & \cellcolor{darkergreen}0.15 & \cellcolor{darkergreen}0.59 \\
\midrule
\multirow{5}{*}{\nlparam} & \multirow{5}{*}{0.660} & \multirow{5}{*}{5} & 
\cellcolor{darkergreen}\hyp{has an element of surprise in its structure} & \cellcolor{darkergreen}0.14 & \cellcolor{darkergreen}0.57 \\
& & & \cellcolor{lightgreen}\hyp{frames the content within a gender-specific context} & \cellcolor{lightgreen}0.11 & \cellcolor{lightgreen}0.53 \\
& & & \cellcolor{darkerred}\hyp{includes an inspiring transformation} & \cellcolor{darkerred}-0.08 & \cellcolor{darkerred}0.51 \\
& & & \cellcolor{lightred}\hyp{uses a play on words or puns} & \cellcolor{lightred}-0.09 & \cellcolor{lightred}0.54 \\
& & & \cellcolor{darkerred}\hyp{poses a rhetorical question} & \cellcolor{darkerred}-0.12 & \cellcolor{darkerred}0.55 \\
\midrule
\multirow{3}{*}{\hypogenic} & \multirow{3}{*}{0.646} & \multirow{3}{*}{3} & 
\cellcolor{darkergreen}\hyp{implies a personal story or anecdote that invites curiosity} & \cellcolor{darkergreen}0.15 & \cellcolor{darkergreen}0.56 \\
& & & \cellcolor{lightgreen}\hyp{uses strong emotional language or evokes curiosity} & \cellcolor{lightgreen}0.12 & \cellcolor{lightgreen}0.58 \\
& & & \cellcolor{darkerred}\hyp{poses a rhetorical question that may disengage the reader} & \cellcolor{darkerred}-0.11 & \cellcolor{darkerred}0.55 \\
\bottomrule
\end{tabular}
\caption{Hypotheses on \headlines with $p < 0.05$ after Bonferroni correction in a multivariate regression. 
`Sep.' is the separation score (\S\ref{sec:theory}): $E[Y | \hat{Z} = 1] - E[Y | \hat{Z} = 0]$, where positive scores predict higher click-rates. The `AUC' column provides AUCs of individual hypotheses. `Overall AUC' uses all 20 hypotheses, including insignificant ones.
Hypotheses for all methods are abbreviated for space.
}
\label{tab:headlines-signif}
\end{table*}

\subsection{Datasets}

We evaluate on synthetic and real-world datasets, described below. Appendix \ref{sec:preprocessing} provides more details on all datasets.

\paragraph{Synthetic datasets: recovering known, reference hypotheses.} 
Our synthetic evaluation is motivated by real-world settings in which there are multiple disjoint hypotheses we would like to discover.
We therefore use two datasets from prior work on interpretable clustering \citep{pham_topicgpt_2024, zhong_explaining_2024}: \textbf{\wiki} and \textbf{\bills}.
We construct a target variable that is positive if a text belongs to one of a pre-specified list of reference categories; these are the ground-truth ``hypotheses'' to recover.
Both datasets are labeled by humans with granular hierarchical topics, such as \textit{Media and Drama: Television: The Simpsons} in \wiki and \textit{Macroeconomics: Tax Code} in \bills. 

We generate ground-truth labels based on the most frequent granular topics: articles in any of the top-5 topics are labeled `1', and all others are labeled `0'.
This produces 5 reference hypotheses, e.g.,~\textit{the bill is about Macroeconomics: Tax Code} is one of them.
For \wiki, we also include a more challenging variation of recovering the top-15 most frequent topics. \wiki contains 11,979 total articles (15\% positive for \wiki-5; 35\% for \wiki-15), and \bills contains 21,149 (24\% positive).
For both datasets, we reserve 2,000 items for validation (i.e.,~SAE hyperparameter selection) and 2,000 heldout items to evaluate hypotheses; we use the remaining items for SAE training and feature selection.

\paragraph{Real-world datasets: Generating hypotheses that predict a target variable.} 
We apply \ourmethod to three real-world datasets which have been studied in prior work:
\begin{itemize}
    \item \textsc{Headlines} \cite{matias_upworthy_2021}: Which features of digital news headlines predict user engagement?
    Each instance is a pair of differently-phrased headlines for the same article; users on Upworthy.com were randomly shown one, and the task is to identify which headline received higher click-rate.
    We reserve a large heldout set to improve statistical power: our split sizes are 8.8K training, 1K  validation, and 4.4K heldout.
    \item \textsc{Yelp} \citep{yelp_yelp_2024}: Which features of Yelp restaurant reviews predict users' 1-5 star ratings? We use 200K reviews for training, 10K for validation, and 10K for heldout eval. 
    \item \textsc{Congress} \citep{gentzkow_what_2010}: Which features of U.S. congressional speeches predict party affiliation?
    Speeches are from the 109th Congress (2005--07) with binary labels (Rep. or Dem.). 
    Our split sizes are 114K training, 16K validation, and 12K heldout.
\end{itemize}

On each of these tasks, our goal is to identify interpretable natural language features that predict the target variable.
The former two are studied in prior work on hypothesis generation \citep{zhou_hypothesis_2024, batista_words_2024, ludan_interpretabledesign_2024}; the latter is a longstanding application in computational social science \citep{grimmer_machine_2021}.

\subsection{Evaluation Metrics}
\label{sec:metrics}

\paragraph{Synthetic datasets.} On the synthetic datasets, we evaluate how well we recover the reference hypotheses.
We run \ourmethod by setting $H$ to the number of references (e.g., 5 for \wiki-5).
We use GPT-4o-mini to annotate each inferred hypothesis on the heldout set, producing an $N_\text{heldout} \times H$ binary matrix of annotations.
Following \citet{zhong_explaining_2024}, we compute the optimal matching between reference and inferred hypotheses via the Hungarian algorithm on the $H \times H$ correlation matrix of reference vs.~inferred hypothesis annotations, and we report three metrics:
\begin{itemize}
    \item \textbf{F1 Similarity}: For each matched (reference, inferred) pair, we compute the F1-score between the presence of the ground-truth reference topic and the inferred hypothesis annotations. 
    We report the mean across the $H$ pairs.
    
    \item \textbf{Surface Similarity}: For each matched pair, we prompt GPT-4o to assess whether the hypotheses are the same, related, or distinct, with corresponding scores of 1.0, 0.5, and 0.0. 
    To improve stability, we sample 5 outputs at temperature 0.7 and average them.
    We report the mean value of this score across the $H$ pairs. (Table \ref{tab:syndata-refvsinferred} shows examples; Figure \ref{fig:surface_similarity_prompt} provides the prompt.)
    
    \item \textbf{Overall AUC}: Using annotations from the $H$ inferred hypotheses, we fit a multivariate logit and compute AUC to evaluate predictive performance. 
    (Recall that the ground-truth label is a logical OR of the reference topics.)
\end{itemize}

The first and second metrics are taken directly from \citet{zhong_explaining_2024}; they assess whether the inferred hypotheses quantitatively and qualitatively match the ground-truth. 
The third metric assesses overall prediction quality. 
Some hypotheses may not match a reference but still be predictive, so this metric rewards any interpretable hypotheses that contribute predictive value, even if they are not optimal.

\paragraph{Real datasets.} On the real datasets, we generate 20 hypotheses per method, and assess the hypothesis sets for two qualities: (1) \textit{breadth}: how many distinct, predictive hypotheses were identified? (2) \textit{predictiveness}: how well do the hypotheses collectively explain the label? 
We again compute an $N_{\text{heldout}} \times H$ annotation matrix for each method's hypotheses, and measure these constructs as follows: 
\begin{itemize}
    \item \textbf{Breadth}: We fit a multivariate logit (for binary labels) or OLS (continuous) regressing the target variable against all of the hypothesis annotations, and report the count of hypotheses that have a significant, nonzero coefficient\footnote{We use a Bonferroni-corrected $p$-value threshold of 2.5e-3.}. 
    This metric benefits distinct hypotheses and penalizes redundancy. 
    Breadth is important since the utility of hypotheses is often determined by criteria beyond predictiveness. 
    For example, a hypothesis may be more useful if it is connected to past theories or suggests a new direction for investigation. More breadth increases the likelihood that a hypothesis satisfying these criteria is produced.
    \item \textbf{Overall Predictive Performance}: Similar to the synthetic datasets, we report the overall prediction quality using all of the hypotheses in a shared regression, reporting AUC for classification and $R^2$ for regression.
    \item \textbf{Qualitative Annotation}: We recruit three computational science researchers to rate each hypothesis for (1) helpfulness and (2) interpretability (details in App.~\ref{sec:qualeval}). These attributes are inspired by prior work \citep{lam2024conceptInduction}.
\end{itemize}

\subsection{Baselines} 
\label{sec:baselines}

\begin{table*}[!htb]
\centering
\fontsize{7.5}{9}\selectfont
\begin{sc}
\begin{tabular}{lccccccccccccccc}
\toprule
& \multicolumn{3}{c}{Wiki, $H=5$} & \multicolumn{3}{c}{Wiki, $H=15$} & \multicolumn{3}{c}{Bills, $H=5$} & \multicolumn{2}{c}{Headlines} & \multicolumn{2}{c}{Yelp} & \multicolumn{2}{c}{Congress} \\
\cmidrule(lr){2-4} \cmidrule(lr){5-7} \cmidrule(lr){8-10} \cmidrule(lr){11-12} \cmidrule(lr){13-14} \cmidrule(lr){15-16}
Method & F1 & Surf. & AUC & F1 & Surf. & AUC & F1 & Surf. & AUC & Ct. & AUC & Ct. & $R^2$ & Ct. & AUC \\
\midrule
\ourmethod & \textbf{0.58} & \textbf{0.70} & \textbf{0.84} & \textbf{0.47} & \textbf{0.61} & \textbf{0.84} & \textbf{0.67} & 0.78 & \textbf{0.89} & \textbf{15} & \textbf{0.69} & \textbf{15} & 0.76 & \textbf{15} & \textbf{0.70} \\
\bertopic & 0.48 & 0.48 & 0.77 & 0.40 & 0.43 & 0.79 & 0.65 & \textbf{0.88} & 0.85 & 1 & 0.62 & 12 & 0.65 & 10 & 0.64 \\
\nlparam & 0.15 & 0.30 & 0.69 & 0.16 & 0.34 & 0.77 & 0.32 & 0.50 & 0.74 & 5 & 0.66 & 11 & 0.58 & 8 & 0.65 \\
\hypogenic & 0.15 & 0.20 & 0.60 & 0.12 & 0.39 & 0.64 & 0.25 & 0.30 & 0.70 & 3 & 0.65 & 12 & \textbf{0.78} & 5 & 0.68 \\
\bottomrule
\end{tabular}
\end{sc}
\caption{\ourmethod performs best at recovering reference human-annotated topics on \wiki-5, \wiki-15, and \bills. We also evaluate on three real-world datasets: \headlines, \yelp, and \congress. ``Ct.'' refers to the count of significant hypotheses, out of 20 candidates; \ourmethod generates the most significant hypotheses.}
\label{tab:metrics-comb}
\vskip -0.1in
\end{table*}

We compare \ourmethod against recent, state-of-the-art methods (\S\ref{sec:relatedwork}). 
For fair comparison, we run all methods with the same embeddings and LLMs as our method, unless stated otherwise:~OpenAI's \embeddingmodel for embeddings; GPT-4o for hypothesis proposal; and GPT-4o-mini for hypothesis annotation.
We run each method to produce $H$ hypotheses and evaluate all methods identically.

\textbf{\bertopic}~\cite{grootendorst_bertopic_2022} is a neural topic model that produces topics by clustering text embeddings.
To produce $H$ hypotheses from \bertopic, texts are clustered into topics, and we fit an $L_1$-regularized model to predict $Y$ from topic ID; like our method, we set the penalty such that there are exactly $H$ nonzero coefficients. 
We assign each topic a natural language label by prompting an LLM with a sample of documents and top terms (prompt in Figure \ref{fig:bertopic_prompt}).

\textbf{\nlparam}~\cite{zhong_explaining_2024} trains a neural network with a length-$K$ bottleneck layer to predict $Y$ from text embeddings, and then uses an LLM to propose labels corresponding to each one of the bottleneck neurons' activations. 
The hypotheses are iteratively refined; a hypothesis is retained only if its annotations (computed on the entire dataset) improve the loss.
We run \nlparam for the default 10 iterations, with 3 candidate hypotheses per neuron. %

\textbf{\hypogenic}~\cite{zhou_hypothesis_2024} directly prompts a language model to propose hypotheses given a task-specific prompt and labeled examples.
They use a bandit reward function to score hypotheses and propose new ones based on incorrectly-labeled examples. 
We run \hypogenic with default parameters; note that the method uses the same model for proposal and annotation, which defaults to GPT-4o-mini.

\textbf{Training set size.}~A key advantage of \ourmethod is that feature generation and selection require only an SAE forward pass, which is cheap even for large datasets; the only LLM-dependent step is feature interpretation.
\bertopic also scales well, as it only uses LLM inference for topic labeling.
In contrast, \nlparam and \hypogenic score and select hypotheses by annotating every training example, so their inference costs scale with training set size. 
Running either method on 200K Yelp reviews, for instance, would cost $>$\$500 and 78 hours (compared to $\sim$\$1 and 20 minutes for \ourmethod). 
To compare to the baselines as favorably as possible, we allow each method to use up to $50\times$ the cost and runtime of ours. 
To fit this budget, we use up to 20,000 training samples for \nlparam and \hypogenic, which accommodates the full training sets on \wiki, \bills, and \headlines; we use random subsets for \yelp and \congress. 
Notably, this sample budget is significantly larger than either method uses in their publication experiments (4,748 for \nlparam; 200 for \hypogenic), suggesting our comparisons are generous to the baselines.

\section{Results}
\label{sec:results}

\subsection{Synthetic Datasets}

\paragraph{\ourmethod recovers ground-truth hypotheses.} 
Table \ref{tab:metrics-comb} shows quantitative metrics: \ourmethod beats all baselines on 8 of 9 dataset-metric pairs. 
As captured by surface similarity, most of \ourmethod's inferred hypotheses either perfectly match or are related to the references.
On \wiki-5 (Table \ref{tab:syndata-refvsinferred}), we retrieve two hypotheses exactly; the other three are slightly too specific (\ourmethod infers a topic \hyp{is about a New York State Route or highway}, but the reference includes all Northeastern roads, not just New York) or broad (it infers \hyp{historical battles} but does not specify \hyp{battles since 1800}), but all inferred hypotheses are always related to a reference, i.e., has surface similarity $\ge$ 0.5. 
No baselines meet this standard on \wiki-5: \nlparam and \hypogenic output 3/5 and 2/5 hypotheses that are unrelated to any of the reference topics, respectively. 
\bertopic performs well, but outputs redundant hypotheses about state highways and misses `battles' entirely.

Our hypotheses also closely match ground-truth when used to annotate heldout data.
The mean F1 score compares how closely annotations according to an inferred hypothesis (e.g.~\hyp{is about a Simpsons episode}) match ground-truth labels (whether the \wiki article belongs to the human-annotated Simpsons category). 
\ourmethod beats all baselines on this metric, especially outperforming \nlparam (+0.36, on average) and \hypogenic (+0.40).
Finally, regressing the label using each method's annotations, \ourmethod achieves higher AUC than all baselines.
Overall, \ourmethod usually outperforms \textsc{BERTopic}—which is designed exactly for this topic inference setting—and substantially outperforms two state-of-the-art LLM hypothesis generation methods.

\subsection{Real-World Datasets}
\label{sec:results_realworld}

Table \ref{tab:metrics-comb} shows quantitative metrics for the real-world datasets. 
Tables \ref{tab:headlines-signif}, \ref{tab:congress-signif}, and \ref{tab:yelp-signif} show all significant hypotheses on \headlines, \congress, and \yelp respectively.

\paragraph{\ourmethod identifies more distinct hypotheses than other methods.} Compared to baselines, \ourmethod produces the most hypotheses which are significant in a multivariate regression.
Importantly, this evaluation reflects successful \textit{generalization}: the hypotheses are learned on the training and validation sets, and they remain significantly associated with the target variable when annotated by GPT-4o-mini on the heldout set.
Across the three datasets, \textbf{45 out of 60 candidate hypotheses are significant}, substantially ahead of the 24, 23, and 20 for \nlparam, \bertopic, and \hypogenic respectively. 

When evaluating overall predictive performance, \ourmethod beats baselines on 8 of 9 comparisons, demonstrating strongly predictive hypotheses. 
The exception is that \hypogenic achieves 2\% higher $R^2$ on Yelp.
However, \hypogenic's most predictive hypotheses, such as \hyp{expresses disappointment with the overall dining experience}, essentially restate the rating prediction task.
While most LLMs can perform such a prediction, these hypotheses do not reveal specific insights about the task. 
In contrast, our hypotheses—in Table \ref{tab:yelp-signif} for \textsc{Yelp}—tend to be more specific: we find that negative reviews mention food poisoning, rude and aggressive staff, or dishonest business practices. 

\begin{figure}[!htb]
\begin{center}
    \includegraphics[width=\linewidth]{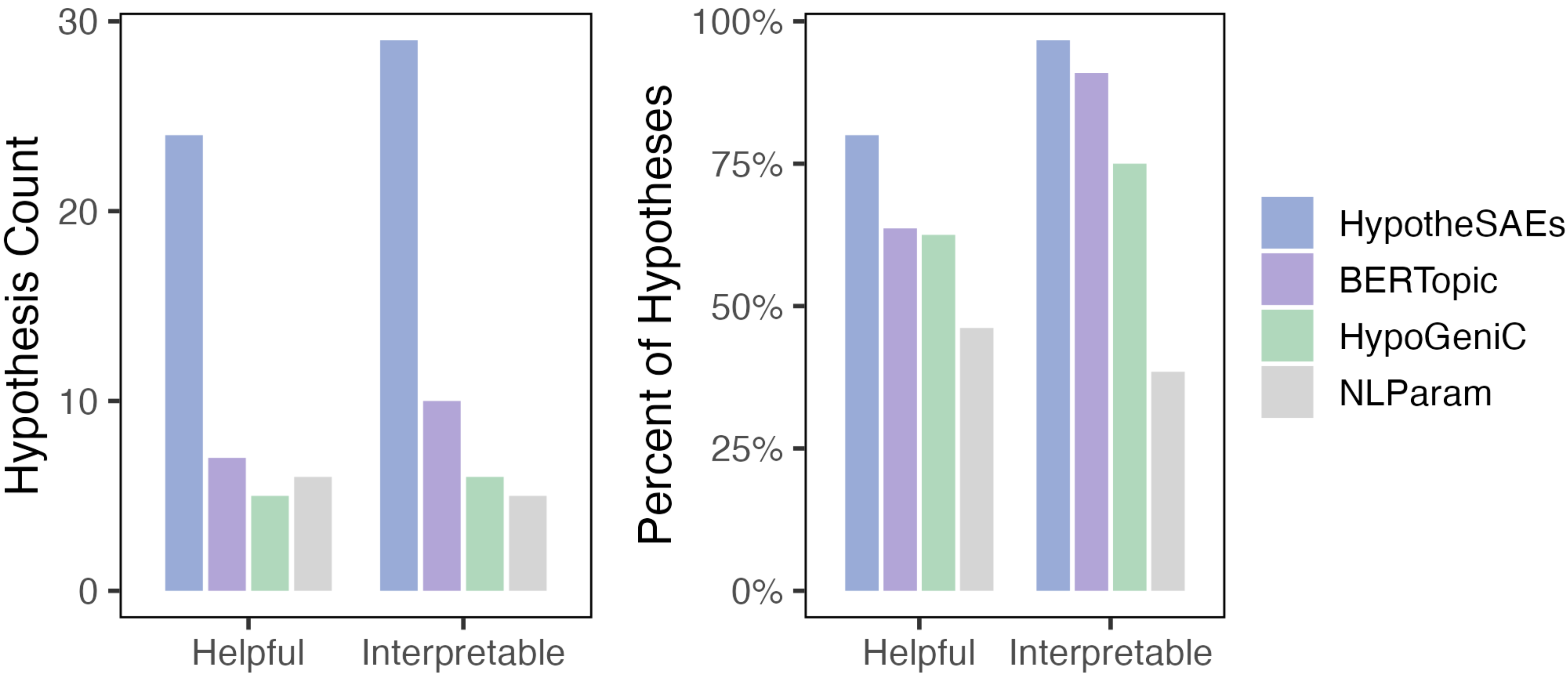}
\end{center}
    \vspace{-1em}
    \caption{\ourmethod produces a larger \textit{number} (left) and \textit{percentage} (right) of significant hypotheses that are rated helpful and interpretable. We report majority vote of 3 ratings per hypothesis.}
    \label{fig:qualeval}
\end{figure}

\paragraph{\ourmethod's outputs are rated as qualitatively helpful and interpretable.} The results of our qualitative evaluation (Figure \ref{fig:qualeval}) confirm the above intuition: most of the predictive concepts generated by \ourmethod are rated helpful (24 out of 30) and interpretable (29 out of 30)\footnote{To reduce annotator workload, our qualitative evaluation only included results from \headlines and \congress, since these are more well-studied social science tasks than \yelp.}. Several hypotheses from \nlparam and \hypogenic, despite being predictive, do not pass these checks.  
The total count of hypotheses that are predictive, helpful, and interpretable generated by \ourmethod (23) far exceeds any baseline (\bertopic is second-best with 7).

\subsection{Non-SAE Ablations}
To isolate the value of the SAE, we conduct ablation experiments. 
Because our method differs from baselines in multiple ways---e.g., our neuron interpretation procedure differs from that of \nlparam---it is worth asking what gain the SAE specifically provides.
To answer this question, we run the following ablations on the real-world datasets:
\begin{itemize}
    \item \textbf{Embedding}: We skip training an SAE and directly select 20 features, via Lasso, from text embeddings. The rest of our method is unchanged.
    \item \textbf{PCA}: Same as \textbf{Embedding}, but we first use PCA to reduce the embedding dimension from 1536 down to 256. 
    \item \textbf{Bottleneck}: Inspired by bottleneck models \citep{yang_language_2023}, we fit a neural network to predict $y$ using text embeddings as input and a single hidden layer with 20 neurons. We generate hypotheses by interpreting the hidden neurons (in the same way we interpret SAE neurons).
\end{itemize}

We show results in Table \ref{tab:ablations}. 
In addition to the count of significant hypotheses and overall predictiveness, we also report average neuron fidelity.

We establish two benefits of the SAE: the neuron interpretations are (1) higher fidelity and (2) cover a wider variety of non-redundant concepts. 
Regarding (1), interpretations of SAE neurons achieve a mean F1 of 0.84, compared to 0.64 for the best ablation (bottleneck).
Regarding (2), we see improvements to downstream metrics: the SAE produces many more significant hypotheses (45/60 across the three datasets, vs.~30/60 for the best ablation) and achieves slightly higher predictiveness.
Qualitatively, the SAE neurons yield more specific interpretations, while the ablations tend to yield generic ones.
Both benefits (1) and (2) are consistent with the SAE producing monosemantic neurons, which are easy to interpret and specific enough to avoid redundancy.

\subsection{Evaluating Hypothesis Novelty}
\label{sec:novelty}

Beyond quantitative comparisons against baselines, we evaluate if \ourmethod discovers new hypotheses even on datasets that have been thoroughly analyzed in prior work.

\paragraph{\congress.} A classic study from \citet{gentzkow_measuring_2016} identifies bigrams and trigrams in Congressional speeches that mark Republican (R) or Democrat (D) speakers. 
We compare to their work by computing counts, in each speech, of their full reported n-gram lists (150 R / 150 D); we also run \ourmethod to produce 100 hypotheses, of which 33 are significant.
A regression using only n-gram counts achieves AUC 0.61, while including our 33 hypotheses increases AUC to 0.74. 
Notably, 28 out of 33 hypotheses remain significantly predictive when controlling for n-grams, showing that our hypotheses add signal to classical methods.

Beyond predictive performance, \ourmethod offers two qualitative advantages. 
First, our hypotheses are inherently interpretable. For example, while \citet{gentzkow_measuring_2016} report that ``oil companies'' correlates with Democrats, it requires context to interpret: are speakers referring to oil companies positively or negatively? In contrast, \ourmethod's corresponding hypothesis is \textit{criticizes oil companies or oil industry practices}. 
Second, our hypotheses capture nuanced patterns that n-grams cannot expresss, like \textit{criticizes government resource allocation, highlighting disparities between domestic needs and actions abroad}.

\paragraph{\headlines.} 
To test our discoveries against more modern methods, we compare to \citet{batista_words_2024}, an LLM hypothesis generation study designed for \headlines and aimed at producing new insights for marketing researchers. 
Each of their 5 hypotheses is validated via human annotation.
Using our heldout evaluation protocol, we compute annotations using their hypotheses, which achieve AUC 0.59 at predicting binary engagement; adding in \ourmethod's 15 significant hypotheses increases AUC to 0.70.
Crucially, \textbf{all 15 of our hypotheses remain significant} when controlling for the prior discoveries in a shared regression.

Several other prior works have also studied \headlines using classical dictionary-based methods \citep{robertson_negativity_2023, gligoric_linguistic_2023, banerjee_language_2024, aubinlequere_when_2025}. 
\ourmethod qualitatively improves on these findings by producing more specific hypotheses.
For example, \citet{robertson_negativity_2023} find that negative words increase engagement, and positive words decrease it.
\ourmethod supports this finding, and also make it more precise: we find that \hyp{situations that evoke discomfort or embarrassment} and mentions of offensive behavior increase clicks, while \hyp{initiatives that positively impact a community} decreases them (Table \ref{tab:headlines-signif}).
\citet{banerjee_language_2024} find that words in an ``emotional intensity'' dictionary increase engagement, which may be difficult to operationalize; our corresponding finding is that headlines referencing \hyp{being surprised, shocked, or unprepared} increase engagement.
Our findings are more precise, and can therefore provide richer insights to researchers.

These results demonstrate that \ourmethod, despite no dataset-specific design, uncovers novel hypotheses that advance prior work from domain experts—suggesting broad applicability across diverse settings.

\subsection{Costs}

In Table \ref{tab:costs}, we report the runtimes, LLM inference token counts, and costs (at current OpenAI API pricing) for all methods on \congress. 
The trends are similar for all datasets.
Runtimes include training the SAEs on one NVIDIA A6000 GPU.
Training is fast because our SAEs can afford to be small ($\sim$$10^3$ features, compared to $10^7$ in \citet{gao_scaling_2024}), since our focus is to learn domain-specific features rather than features of all language.

\ourmethod is much more efficient than \nlparam and \hypogenic. 
Despite the fact that these methods are trained on less data---20K examples vs.~114K for \ourmethod---\textbf{\ourmethod is $\sim$30-50$\times$ faster and 10-50$\times$ cheaper}.
This is because \nlparam and \hypogenic score hypotheses by annotating every example in the dataset. 
As a result, their LLM annotation requirements scale with both the number of hypotheses $H$ and dataset size $N$, while \ourmethod's scales only with $H$.

\begin{table}[H]
\centering
\renewcommand{\arraystretch}{1.1}
\small
\begin{tabular}{lrrr}
\toprule
Method & Time (min) & Tokens (M) & Cost \\
\midrule
\ourmethod & 18.1 & 0.2 / 7.2 & \$1.29 \\
\textsc{Sae-NoVal} & 7.9 & 0.1 / 0.0 & \$0.15 \\
\nlparam & 904.9 & 2.8 / 414 & \$69.92 \\
\hypogenic & 575.4 & 0.0 / 72 & \$11.02 \\
\bertopic & 3.0 & 0.1 / 0.0 & \$0.19 \\
\bottomrule
\end{tabular}
\caption{
Runtimes and costs for hypothesis generation on \congress. 
We report millions of tokens for GPT-4o (concept generation) and GPT-4o-mini (concept annotation) respectively as \{\# 4o\} / \{\# 4o-mini\}. Input and output token counts are summed ($>$99\% are input tokens).
\textsc{Sae-NoVal} is our method without the fidelity validation step (i.e., for each neuron, we generate only one label instead of choosing from three based on fidelity).
Costs are in USD and based on OpenAI pricing as of 01/2025.
}
\label{tab:costs}
\end{table}

\bertopic is cheapest, and remains a good baseline for a resource-limited analysis.
However, \ourmethod can also be made cheaper by skipping the non-essential label validation step: instead of generating 3 candidate labels per neuron and choosing the highest-fidelity one, we can simply generate 1 label per neuron and use it directly.
This reduces cost to that of \bertopic, while still exceeding its performance; we describe this ablation in Appendix \ref{sec:hyperparams_neuronlabeling}.

\section{Conclusion}

We propose \ourmethod, a scalable method to generate interpretable hypotheses from black-box text representations.
On three synthetic and three real-world tasks, \ourmethod outperforms state-of-the-art topic modeling and LLM baselines, at less than 10$\times$ the time and cost of the latter. 
Beyond the tasks we study, the method applies naturally to the many settings in the social sciences where a large text corpus is labeled with a target variable of interest \citep{ziems_can_2024, bollen_twitter_2011, card_computational_2022, breuer_using_2025}.
In future work, an exciting direction is extending our method to explain non-text datasets, like images \citep{dunlap_describing_2024} and proteins \citep{vig_bertology_2021}, as well as text datasets from specialized domains like healthcare \citep{hsu_clinical_2023, robitschek_large_2025}.
More generally, there are many applications where we can accurately \textit{predict} a target variable, but producing new scientific insights remains difficult. Our work helps bridge this gap.

\section*{Software and Data}
Code is \href{https://github.com/rmovva/HypotheSAEs}{available on GitHub \faicon{github}}, and can be installed via pip with \texttt{pip install hypothesaes}. 
A notebook to reproduce experimental results in the paper is available in the repository.
All data used in the paper are \href{https://huggingface.co/datasets/rmovva/HypotheSAEs}{available on HuggingFace}.
A demo to visualize all SAE neurons and how well they predict the target for \headlines, \yelp, and \congress is available at \,\href{https://hypothesaes.org/}{\faicon{globe} \texttt{hypothesaes.org}}.

\section*{Acknowledgements}
Thanks to Rishi Jha, Hal Triedman, Anna Lyubarskaja, Ruiqi Zhong, Serina Chang, and Sammi Cheung for helpful discussions. 
Thanks to Sidhika Balachandar, Divya Shanmugam, and Gabriel Agostini for qualitative annotation.
Thanks to the ICML reviewers for helpful suggestions (specifically, to include SAE ablations and qualitative evaluation).
Thanks to Neil Movva and OpenAI for LLM inference credits.

\textbf{Funding:} RM is supported by NSF DGE \#2146752.  
NG is supported by NSF CAREER \#2339427, and Cornell Tech Urban Tech Hub, Meta, and Amazon research awards. 
JK is supported by a Simons Collaboration grant and a grant from the MacArthur Foundation.
EP is supported by a Google Research Scholar award, an AI2050 Early Career Fellowship, NSF CAREER \#2142419, a CIFAR Azrieli Global scholarship, a gift to the LinkedIn-Cornell Bowers CIS Strategic Partnership, the Survival and Flourishing Fund, Open Philanthropy, and the Abby Joseph Cohen Faculty Fund.

\section*{Impact Statement}

Our work advances the field of AI-assisted hypothesis generation, which we hope will drive progress in the social and natural sciences.
To the best of our knowledge, there are no particular negative social consequences imposed by our work compared to machine learning research in general.

\bibliography{raj,kenny}
\bibliographystyle{icml2025}

\newpage
\appendix
\onecolumn
\newlength{\ultralightrulewidth}
\setlength{\ultralightrulewidth}{0.02em}  %
\setlength{\arrayrulewidth}{0.02em}
\begin{table*}[!htbp]
\vskip \captionvskip
\begin{center}
\scriptsize 
{\setlength{\arrayrulewidth}{0.4pt}  %
\begin{tabular}{p{6cm}|p{5cm}cc}
\toprule
\textsc{Reference} & \textsc{Inferred} & \textsc{F1} & \textsc{Surface} \\ 
\midrule
Other music articles $>$ Performers, groups, composers, and other music-related people & \textit{is about a rock band or a musical act} & 0.39 & 0.5 \\
\midrule
Earth science $>$ Tropical cyclones: Atlantic & \textit{is about hurricanes or tropical storms} & 0.52 & 1.0 \\
\midrule
Battles, exercises, and conflicts $>$ Modern history (1800 to present) & \textit{is about a historical battle} & 0.36 & 0.5 \\
\midrule
Television $>$ The Simpsons episodes & \textit{is about episodes of The Simpsons television show} & 0.92 & 1.0 \\
\midrule
Transport $>$ Road infrastructure: Northeastern United States & \textit{is about a New York State Route} & 0.71 & 0.5 \\
\bottomrule
\end{tabular}}
\caption{Evaluating hypotheses inferred by \ourmethod against the reference topics on \wiki.}
\label{tab:syndata-refvsinferred}
\end{center}
\end{table*}

\begin{table*}[!htbp]
\centering
\scriptsize
\renewcommand{\arraystretch}{1.2}
\definecolor{lightred}{HTML}{ffe3e3}
\definecolor{darkerred}{HTML}{ffd1d1}
\definecolor{lightblue}{HTML}{e3f2fd}
\definecolor{darkerblue}{HTML}{bbdefb}

\begin{tabular}{lccp{8cm}rr}
\toprule
Source & Overall AUC & \# Sig & Significant Hypotheses: \colorbox{darkerred}{Red: $\uparrow$ Republican}; \colorbox{darkerblue}{Blue: $\uparrow$ Democrat} & Sep. & AUC \\
\midrule
\multirow{15}{*}{\ourmethod}& \multirow{15}{*}{0.702} & \multirow{15}{*}{15} & 
\cellcolor{lightred}\textit{contains the phrase 'I ask unanimous consent'} & \cellcolor{lightred}0.30 & \cellcolor{lightred}0.53 \\
& & & \cellcolor{darkerred}\textit{mentions hearings conducted by Senate Committees or Subcommittees} & \cellcolor{darkerred}0.18 & \cellcolor{darkerred}0.52 \\
& & & \cellcolor{lightred}\textit{mentions scheduling or details about Senate votes or amendments} & \cellcolor{lightred}0.17 & \cellcolor{lightred}0.55 \\
& & & \cellcolor{darkerred}\textit{discusses illegal immigration and its associated implications} & \cellcolor{darkerred}0.15 & \cellcolor{darkerred}0.51 \\
& & & \cellcolor{lightred}\textit{mentions victories or successes in military or political contexts} & \cellcolor{lightred}0.15 & \cellcolor{lightred}0.51 \\
& & & \cellcolor{darkerred}\textit{mentions freedom or liberty} & \cellcolor{darkerred}0.06 & \cellcolor{darkerred}0.50 \\
& & & \cellcolor{lightred}\textit{discusses economic growth or growth rates} & \cellcolor{lightred}0.01 & \cellcolor{lightred}0.50 \\
& & & \cellcolor{lightblue}\textit{discusses government spending or budgetary issues} & \cellcolor{lightblue}-0.09 & \cellcolor{lightblue}0.54 \\
& & & \cellcolor{darkerblue}\textit{mentions the need for reform or proposes reforms} & \cellcolor{darkerblue}-0.17 & \cellcolor{darkerblue}0.58 \\
& & & \cellcolor{lightblue}\textit{mentions key figures related to the civil rights movement} & \cellcolor{lightblue}-0.22 & \cellcolor{lightblue}0.51 \\
& & & \cellcolor{darkerblue}\textit{criticizes government policies or actions} & \cellcolor{darkerblue}-0.29 & \cellcolor{darkerblue}0.63 \\
& & & \cellcolor{lightblue}\textit{mentions the U.S. national debt or raising the debt limit} & \cellcolor{lightblue}-0.40 & \cellcolor{lightblue}0.51 \\
& & & \cellcolor{darkerblue}\textit{criticizes tax breaks or advantages for the wealthy} & \cellcolor{darkerblue}-0.44 & \cellcolor{darkerblue}0.52 \\
& & & \cellcolor{lightblue}\textit{criticizes Republican leadership or policies} & \cellcolor{lightblue}-0.45 & \cellcolor{lightblue}0.60 \\
& & & \cellcolor{darkerblue}\textit{criticizes the administration's handling of the Iraq war} & \cellcolor{darkerblue}-0.45 & \cellcolor{darkerblue}0.52 \\
\midrule
\multirow{10}{*}{\bertopic} & \multirow{10}{*}{0.636} & \multirow{10}{*}{10} & 
\cellcolor{lightred}\textit{Senate procedural requests and adjournment schedules} & \cellcolor{lightred}0.39 & \cellcolor{lightred}0.53 \\
& & & \cellcolor{darkerred}\textit{Requests for unanimous consent to authorize committee meetings} & \cellcolor{darkerred}0.32 & \cellcolor{darkerred}0.53 \\
& & & \cellcolor{lightred}\textit{procedural discussions and scheduling of votes} & \cellcolor{lightred}0.20 & \cellcolor{lightred}0.56 \\
& & & \cellcolor{darkerred}\textit{discusses immigration and border security} & \cellcolor{darkerred}0.13 & \cellcolor{darkerred}0.51 \\
& & & \cellcolor{lightred}\textit{mentions individuals with professional titles or roles} & \cellcolor{lightred}0.08 & \cellcolor{lightred}0.53 \\
& & & \cellcolor{darkerred}\textit{debates on earmarks and federal spending processes} & \cellcolor{darkerred}0.05 & \cellcolor{darkerred}0.50 \\
& & & \cellcolor{lightblue}\textit{discusses the Darfur conflict and international intervention} & \cellcolor{lightblue}-0.30 & \cellcolor{lightblue}0.50 \\
& & & \cellcolor{darkerblue}\textit{Congressional oversight of defense contracting in Iraq} & \cellcolor{darkerblue}-0.33 & \cellcolor{darkerblue}0.50 \\
& & & \cellcolor{lightblue}\textit{discusses Asian Pacific American communities} & \cellcolor{lightblue}-0.36 & \cellcolor{lightblue}0.50 \\
& & & \cellcolor{darkerblue}\textit{Critiques Republican majority leadership} & \cellcolor{darkerblue}-0.46 & \cellcolor{darkerblue}0.55 \\
\midrule
\multirow{8}{*}{\nlparam} & \multirow{8}{*}{0.650} & \multirow{8}{*}{8} & 
\cellcolor{lightred}\textit{expresses strong patriotism} & \cellcolor{lightred}0.09 & \cellcolor{lightred}0.51 \\
& & & \cellcolor{darkerred}\textit{contains specific legislative references} & \cellcolor{darkerred}0.08 & \cellcolor{darkerred}0.53 \\
& & & \cellcolor{lightblue}\textit{mentions specific individuals or organizations} & \cellcolor{lightblue}-0.03 & \cellcolor{lightblue}0.52 \\
& & & \cellcolor{darkerblue}\textit{includes numerical data} & \cellcolor{darkerblue}-0.08 & \cellcolor{darkerblue}0.53 \\
& & & \cellcolor{lightblue}\textit{addresses a critical national issue} & \cellcolor{lightblue}-0.14 & \cellcolor{lightblue}0.56 \\
& & & \cellcolor{darkerblue}\textit{expresses strong support or opposition} & \cellcolor{darkerblue}-0.15 & \cellcolor{darkerblue}0.58 \\
& & & \cellcolor{lightblue}\textit{advocates for specific groups or communities} & \cellcolor{lightblue}-0.17 & \cellcolor{lightblue}0.55 \\
& & & \cellcolor{darkerblue}\textit{employs emotional or dramatic language} & \cellcolor{darkerblue}-0.23 & \cellcolor{darkerblue}0.58 \\
\midrule
\multirow{5}{*}{\hypogenic} & \multirow{5}{*}{0.675} & \multirow{5}{*}{5} & 
\cellcolor{lightred}\textit{focuses on national security and immigration enforcement} & \cellcolor{lightred}0.18 & \cellcolor{lightred}0.51 \\
& & & \cellcolor{darkerblue}\textit{discusses social issues such as healthcare and civil rights} & \cellcolor{darkerblue}-0.22 & \cellcolor{darkerblue}0.58 \\
& & & \cellcolor{lightblue}\textit{mentions of government inefficiencies or calls for reform} & \cellcolor{lightblue}-0.27 & \cellcolor{lightblue}0.62 \\
& & & \cellcolor{darkerblue}\textit{advocates for civil liberties and critiques government overreach} & \cellcolor{darkerblue}-0.29 & \cellcolor{darkerblue}0.57 \\
& & & \cellcolor{lightblue}\textit{emphasizes the need for government accountability} & \cellcolor{lightblue}-0.37 & \cellcolor{lightblue}0.62 \\
\bottomrule
\end{tabular}
\caption{\congress hypotheses, for each method, with $p< 0.05$ after Bonferroni correction in a multivariate regression. The `Sep.' column shows the separation score: $E[Y | \hat{Z} = 1] - E[Y | \hat{Z} = 0]$, where positive scores predict Republican speech and negative scores predict Democratic speech. The `AUC' column provides univariate AUCs for individual hypotheses. `Overall AUC' uses all hypotheses, including insignificant ones.
}
\label{tab:congress-signif}
\end{table*}

\begin{table*}[!htbp]
\centering
\scriptsize
\renewcommand{\arraystretch}{1.2}
\definecolor{darkergreen}{HTML}{c8e6c9}
\definecolor{lightgreen}{HTML}{e8f5e9}
\definecolor{darkerred}{HTML}{ffd1d1}
\definecolor{lightred}{HTML}{ffe3e3}

\begin{tabular}{lccp{8cm}rr}
\toprule
Source & Overall $R^2$ & \# Sig & Significant Hypotheses: \colorbox{darkergreen}{Green: $\uparrow$ rating}; \colorbox{darkerred}{Red: $\downarrow$ rating} & Sep. & $R^2$ \\
\midrule
\multirow{15}{*}{\ourmethod}& \multirow{15}{*}{0.755} & \multirow{15}{*}{15} & 
\cellcolor{darkergreen}\textit{expresses extreme enthusiasm or excitement with phrases like 'holy fuck', 'can't wait to go back', or extensive use of exclamation points} & \cellcolor{darkergreen}1.45 & \cellcolor{darkergreen}0.23 \\
& & & \cellcolor{lightgreen}\textit{uses exclamation marks enthusiastically to convey positive sentiment} & \cellcolor{lightgreen}1.41 & \cellcolor{lightgreen}0.23 \\
& & & \cellcolor{darkergreen}\textit{mentions anticipation or plans to return to the restaurant or order again} & \cellcolor{darkergreen}1.20 & \cellcolor{darkergreen}0.17 \\
& & & \cellcolor{lightgreen}\textit{explicitly states that a food item or restaurant is the best, often using superlative language} & \cellcolor{lightgreen}1.10 & \cellcolor{lightgreen}0.07 \\
& & & \cellcolor{darkerred}\textit{mentions mediocrity or averageness of food or experience} & \cellcolor{darkerred}-1.84 & \cellcolor{darkerred}0.31 \\
& & & \cellcolor{lightred}\textit{mentions issues related to food safety, hygiene, or improper food handling} & \cellcolor{lightred}-1.98 & \cellcolor{lightred}0.15 \\
& & & \cellcolor{darkerred}\textit{mentions issues with restaurant staff communication or behavior} & \cellcolor{darkerred}-2.23 & \cellcolor{darkerred}0.44 \\
& & & \cellcolor{lightred}\textit{mentions dissatisfaction or criticism of service, staff, or management} & \cellcolor{lightred}-2.25 & \cellcolor{lightred}0.61 \\
& & & \cellcolor{darkerred}\textit{mentions food poisoning or symptoms of foodborne illness} & \cellcolor{darkerred}-2.32 & \cellcolor{darkerred}0.02 \\
& & & \cellcolor{lightred}\textit{mentions rude or aggressive behavior from restaurant staff} & \cellcolor{lightred}-2.37 & \cellcolor{lightred}0.30 \\
& & & \cellcolor{darkerred}\textit{mentions dissatisfaction with restaurant policies, rules, or restrictions} & \cellcolor{darkerred}-2.37 & \cellcolor{darkerred}0.53 \\
& & & \cellcolor{lightred}\textit{criticizes the behavior or attitude of staff} & \cellcolor{lightred}-2.40 & \cellcolor{lightred}0.50 \\
& & & \cellcolor{darkerred}\textit{complains about rude or unprofessional behavior from staff or management} & \cellcolor{darkerred}-2.42 & \cellcolor{darkerred}0.46 \\
& & & \cellcolor{lightred}\textit{mentions dishonest or unethical business practices by the establishment} & \cellcolor{lightred}-2.48 & \cellcolor{lightred}0.17 \\
& & & \cellcolor{darkerred}\textit{describes a specific instance of poor customer service} & \cellcolor{darkerred}-2.50 & \cellcolor{darkerred}0.50 \\
\midrule
\multirow{12}{*}{\bertopic} & \multirow{12}{*}{0.647} & \multirow{12}{*}{12} & 
\cellcolor{darkergreen}\textit{praises the combination of excellent food quality and friendly, attentive service} & \cellcolor{darkergreen}1.61 & \cellcolor{darkergreen}0.33 \\
& & & \cellcolor{lightgreen}\textit{describes detailed experiences with food quality, preparation, and flavor nuances} & \cellcolor{lightgreen}0.46 & \cellcolor{lightgreen}0.03 \\
& & & \cellcolor{darkerred}\textit{criticizes the freshness, quality, or preparation of sushi and fish dishes} & \cellcolor{darkerred}-1.77 & \cellcolor{darkerred}0.07 \\
& & & \cellcolor{lightred}\textit{describes issues with restaurant closures or inaccurate operating hours} & \cellcolor{lightred}-1.77 & \cellcolor{lightred}0.10 \\
& & & \cellcolor{darkerred}\textit{complains about uncleanliness of restaurant spaces} & \cellcolor{darkerred}-1.91 & \cellcolor{darkerred}0.05 \\
& & & \cellcolor{lightred}\textit{criticizes the quality and authenticity of Chinese dishes} & \cellcolor{lightred}-1.94 & \cellcolor{lightred}0.06 \\
& & & \cellcolor{darkerred}\textit{criticizes Italian dishes, particularly pasta and sauces} & \cellcolor{darkerred}-1.94 & \cellcolor{darkerred}0.04 \\
& & & \cellcolor{lightred}\textit{complains about issues with pizza orders} & \cellcolor{lightred}-2.08 & \cellcolor{lightred}0.16 \\
& & & \cellcolor{darkerred}\textit{criticizes breakfast food quality, portion sizes, and service experience} & \cellcolor{darkerred}-2.15 & \cellcolor{darkerred}0.37 \\
& & & \cellcolor{lightred}\textit{complains about issues with delivery orders} & \cellcolor{lightred}-2.20 & \cellcolor{lightred}0.25 \\
& & & \cellcolor{darkerred}\textit{complains about poor service related to waiting times and staff behavior} & \cellcolor{darkerred}-2.31 & \cellcolor{darkerred}0.49 \\
& & & \cellcolor{lightred}\textit{complains about rude or unprofessional behavior from bartenders or bar staff} & \cellcolor{lightred}-2.31 & \cellcolor{lightred}0.31 \\
\midrule
\multirow{11}{*}{\nlparam} & \multirow{11}{*}{0.583} & \multirow{11}{*}{11} & 
\cellcolor{darkergreen}\textit{uses enthusiastic or excessively positive language} & \cellcolor{darkergreen}1.90 & \cellcolor{darkergreen}0.47 \\
& & & \cellcolor{lightgreen}\textit{describes exceptional food quality} & \cellcolor{lightgreen}1.65 & \cellcolor{lightgreen}0.35 \\
& & & \cellcolor{darkergreen}\textit{describes a positive recurring experience} & \cellcolor{darkergreen}1.45 & \cellcolor{darkergreen}0.26 \\
& & & \cellcolor{lightgreen}\textit{recommends the restaurant to others} & \cellcolor{lightgreen}1.29 & \cellcolor{lightgreen}0.18 \\
& & & \cellcolor{darkergreen}\textit{uses exclamation marks to express enthusiasm} & \cellcolor{darkergreen}1.22 & \cellcolor{darkergreen}0.17 \\
& & & \cellcolor{lightgreen}\textit{mentions attentive or friendly staff} & \cellcolor{lightgreen}1.01 & \cellcolor{lightgreen}0.12 \\
& & & \cellcolor{darkergreen}\textit{expresses enthusiasm for the ambiance or atmosphere} & \cellcolor{darkergreen}0.92 & \cellcolor{darkergreen}0.07 \\
& & & \cellcolor{lightgreen}\textit{mentions thoughtful accommodation of customer preferences} & \cellcolor{lightgreen}0.79 & \cellcolor{lightgreen}0.01 \\
& & & \cellcolor{darkergreen}\textit{describes generosity in portion sizes} & \cellcolor{darkergreen}0.69 & \cellcolor{darkergreen}0.02 \\
& & & \cellcolor{lightgreen}\textit{comments on portion size} & \cellcolor{lightgreen}0.02 & \cellcolor{lightgreen}0.00 \\
& & & \cellcolor{darkerred}\textit{mentions long wait times to receive service} & \cellcolor{darkerred}-1.70 & \cellcolor{darkerred}0.15 \\
\midrule
\multirow{12}{*}{\hypogenic} & \multirow{12}{*}{0.775} & \multirow{12}{*}{12} & 
\cellcolor{darkergreen}\textit{highlights a pleasant atmosphere or enjoyable dining experience} & \cellcolor{darkergreen}2.40 & \cellcolor{darkergreen}0.66 \\
& & & \cellcolor{lightgreen}\textit{expresses a willingness to return or recommend the restaurant to others} & \cellcolor{lightgreen}2.13 & \cellcolor{lightgreen}0.55 \\
& & & \cellcolor{darkergreen}\textit{expresses a desire to return or recommends the place to others} & \cellcolor{darkergreen}1.89 & \cellcolor{darkergreen}0.45 \\
& & & \cellcolor{lightgreen}\textit{describes food as delicious or highlights specific dishes with positive adjectives} & \cellcolor{lightgreen}1.83 & \cellcolor{lightgreen}0.42 \\
& & & \cellcolor{darkerred}\textit{notes issues with food quality, such as being cold or poorly cooked} & \cellcolor{darkerred}-1.87 & \cellcolor{darkerred}0.29 \\
& & & \cellcolor{lightred}\textit{includes complaints about long wait times for food or service} & \cellcolor{lightred}-1.94 & \cellcolor{lightred}0.30 \\
& & & \cellcolor{darkerred}\textit{expresses disappointment with the quality of food compared to expectations} & \cellcolor{darkerred}-2.17 & \cellcolor{darkerred}0.49 \\
& & & \cellcolor{lightred}\textit{mentions poor service or unresponsive staff during the dining experience} & \cellcolor{lightred}-2.36 & \cellcolor{lightred}0.51 \\
& & & \cellcolor{darkerred}\textit{complains about poor customer service or rude staff} & \cellcolor{darkerred}-2.41 & \cellcolor{darkerred}0.50 \\
& & & \cellcolor{lightred}\textit{complains about poor service or unhelpful staff} & \cellcolor{lightred}-2.42 & \cellcolor{lightred}0.60 \\
& & & \cellcolor{darkerred}\textit{expresses disappointment with the overall dining experience} & \cellcolor{darkerred}-2.42 & \cellcolor{darkerred}0.67 \\
& & & \cellcolor{lightred}\textit{expresses disappointment with the overall experience or ambiance} & \cellcolor{lightred}-2.43 & \cellcolor{lightred}0.67 \\
\bottomrule
\end{tabular}
\caption{\yelp hypotheses, for each method, with $p < 0.05$ after Bonferroni correction in a multivariate regression. 
`Sep.' is the separation score: $E[Y | \hat{Z} = 1] - E[Y | \hat{Z} = 0]$, where positive scores predict higher restaurant ratings. The `$R^2$' column provides univariate $R^2$ values of individual hypotheses. `Overall $R^2$' uses all 20 hypotheses, including insignificant ones.
Hypotheses for all methods are abbreviated.
}
\label{tab:yelp-signif}
\end{table*}

\begin{table}[htbp]
\centering
\renewcommand{\arraystretch}{1.2}
\small
\begin{tabular}{lrrr}
\toprule
Method & Total Count & Avg. Predictiveness & Avg. Fidelity \\
\midrule
SAE & \textbf{45 / 60} & \textbf{0.717} & \textbf{0.836} \\
Embedding & 30 / 60 & 0.706 & 0.545 \\
PCA & 23 / 60 & 0.697 & 0.547 \\
Bottleneck & 27 / 60 & 0.699 & 0.640 \\
\bottomrule
\end{tabular}
\caption{
Experiments testing ablations of the SAE on the three real-world datasets: \headlines, \yelp, \congress. 
Using the SAE yields many more significant hypotheses, slightly more predictive hypotheses, and much higher fidelity interpretations (i.e., interpretations that better explain the underlying neuron activations).
The count of significant hypotheses is aggregated across the three datasets.
Predictiveness is averaged across the three datasets.
Fidelity is averaged across the 60 selected neurons (3 datasets, 20 neurons selected per dataset).
}
\label{tab:ablations}
\end{table}

\section{Data Preprocessing}
\label{sec:preprocessing}

\paragraph{\wiki.} 
We download the dataset as processed by \citet{pham_topicgpt_2024}; the data are derived from the WikiText corpus assembled by \citet{merity_pointer_2016}. 
The articles are categorized by Wikipedia editors; each article has a supercategory, category, subcategory (e.g., \smalltt{Media and Drama > Television > The Simpsons Episodes}). 
We focus on recovering article subcategories, since these are the most specific and therefore most difficult to fully recover.
After filtering out duplicates and infrequent ($<$100 articles) subtopics, the dataset contains 11,979 items spanning 69 subcategories.
For the \wiki-5 dataset, we set the ground-truth topics to the 5 most common subcategories. 
That is, we label all 1,771 articles (14.8\%) in these subcategories as positives, while all other article subcategories are negatives.
The \wiki-15 dataset is constructed similarly using the 15 most common subcategories (4,190 positives, 35.0\%).

\paragraph{\bills.}
We also download \bills from \citet{pham_topicgpt_2024}; the dataset was originally assembled by \citet{hoyle_are_2022} using GovTrack\footnote{\href{https://www.govtrack.us/congress/bills/}{https://www.govtrack.us/congress/bills/}}.
The bills are from the 110th-114th U.S. Congresses (Jan 2007 - Jan 2017).
Each bill has a topic and subtopic.
After filtering out very short bill texts (which may not have been properly transcribed), duplicate bill texts, bills with no subtopic, and bills with infrequent ($<$100) subtopics, there are 20,834 items spanning 70 subtopics.
We label the 5,038 bills (24.2\%) with the 5 most common subtopics as positives, and all others as negatives.

\paragraph{\headlines.} We use the \headlines dataset as collected and released by the Upworthy Research Archive \citep{matias_upworthy_2021}.
All of the data is derived from web traffic to Upworthy.com, a digital media platform known for its attention-grabbing articles.
The dataset consists of thousands of A/B tests, where users were randomly shown one of several possible headline variations for the same underlying article.
Each headline is linked to a corresponding number of impressions and clicks, allowing researchers to estimate the causal effect of headline text on \textit{click-rate} (clicks$/$impressions, also referred to as clickthrough-rate or engagement).

To preprocess the data, we start by downloading all publicly available data from the Upworthy archive\footnote{\href{https://upworthy.natematias.com/index}{https://upworthy.natematias.com/index}}, which includes clicks and impressions for 122,565 distinct headlines (we remove 28K rows from a problematic data range as recently reported by the dataset authors\footnote{\href{https://upworthy.natematias.com/2024-06-upworthy-archive-update.html}{https://upworthy.natematias.com/2024-06-upworthy-archive-update.html}}). 
We then group headlines which have the same \smalltt{test\_id} (signifying that those headlines were randomized against one another).
There are 14,128 such groups with at least two headlines.
Following \citet{zhou_hypothesis_2024}, we take the headlines with max and min click-rates in each group, and design a binary classification task to predict which headline in the pair received higher engagement.
That is, we construct tuples (headline A, headline B, label), where headline A and headline B are randomly shuffled, and the label is 1 if headline A received higher click-rate than headline B.

Following \citet{batista_words_2024}, we take specific care to prevent train-holdout leakage on this dataset. 
Specifically, the same headline texts may be used in multiple randomized experiments (i.e., the same headline may occur in different \smalltt{test\_id}'s).
We ensure that no identical headlines will appear in both the train split and the heldout split; we encourage future researchers to follow this practice, since splitting on \smalltt{test\_id} alone does not fully prevent headline leakage.

To handle the pairwise design, we train the SAE on the embeddings of all unique headlines in the training set, and then compute the activation matrices $Z_{\text{SAE}}^A$ and $Z_{\text{SAE}}^B$ for all of the headline A's and headline B's respectively.
We select predictive neurons by identifying columns in the matrix difference $\Delta Z_{\text{SAE}} := Z_{\text{SAE}}^A - Z_{\text{SAE}}^B$ which are predictive of $\mathbf{y}$.

\paragraph{\yelp.} We download the Yelp Open Dataset\footnote{\href{https://business.yelp.com/data/resources/open-dataset/}{https://business.yelp.com/data/resources/open-dataset/}}, and we filter to the 4.72M reviews where the business contains the `Restaurant' tag.
From this, we randomly sample training, validation, and heldout sets (200K, 10K, and 10K respectively).

\paragraph{\congress.} 
We download all congressional speech transcripts from the 109th U.S. Congress \citep{gentzkow_what_2010}\footnote{\href{https://data.stanford.edu/congress_text}{https://data.stanford.edu/congress\_text}}, which met from January 2005 to January 2007. 
We filter speeches which are very short (less than 250 characters) and include only speeches from Republican or Democrat speakers; the dataset is roughly balanced (51.7\% Republican).
If a speech is longer than ten sentences, we split it into ten sentence chunks, and exclude any chunks beyond the first five (to avoid over-representation of specific speeches).
For the heldout set, we restrict to one chunk per speech to ensure that our hypotheses generalize to a diverse corpus. 
Ultimately, we use 114K speech chunks for training, 16K for validation, and 12K for heldout eval. 
If a speech has multiple excerpts, we ensure that all of them are included in the same split to avoid leakage.

\section{Hyperparameters}
\label{sec:hyperparams}

\subsection{Training sparse autoencoders} 
We train $k$-sparse autoencoders, where each forward pass masks all neurons besides the top-$k$ activating ones \citep{makhzani_ksparse_2014, gao_scaling_2024}.
The most influential hyperparameters, which we tune for each dataset, are $M$ and $k$.
We choose $(M, k)$ by maximizing validation performance. 
Specifically, for each combination of $(M, k)$, we evaluate validation AUC (or $R^2$ on \yelp) after fitting an $L_1$-regularized predictor with exactly $H \ll M$ nonzero coefficients. 
We constrain $M$ and $k$ to powers of two to tractably limit the search space.

These parameters, broadly, control the level of granularity of the concepts learned by each SAE neuron: $M$ is the total number of concepts which can be learned across the entire dataset, and $k$ is the number of concepts which can be used to represent each instance.
For intuition, assume that with $(M, k) = (16, 4)$ on \yelp, the SAE learns a single neuron that fires when reviews mention \textit{price}; with $(M, k) = (32, 4)$, the SAE may learn separate neurons for reviews which mention \textit{high prices} vs.~\textit{low prices} (described as ``feature splitting'' in \citet{bricken2023towards}); with $(M, k) = (32, 8)$, the SAE may learn to represent more niche features altogether, such as the \textit{number of exclamation marks}.
In general, increasing either parameter yields more granular features, at the risk of producing some neurons that are uninterpretable or redundant.

In some cases we would like features at multiple levels of granularity: for example, we may want both the general feature ``mentions prices'' and the more specific feature ``mentions that the cocktails were cheap during happy hour'', but training a single large SAE (e.g. $(1024, 32)$) may only produce features as specific as the latter. 
A simple solution is to train multiple SAEs of different sizes, concatenate their activation matrices, and select features from any of them. 
We find that this approach is helpful (improves validation AUC) on the \headlines dataset. 
Ultimately, we use the following values:
 
\begin{itemize}
    \item \wiki-5: $(32, 4)$.
    \item \wiki-15 \& \bills: $(64, 4)$.
    \item \headlines: $(256, 8); (32, 4)$.
    \item \yelp: $(1024, 32)$.
    \item \congress: $(4096, 32)$.
\end{itemize}

Following prior work, we add several optimizations to improve SAE training. Following best practices based on results from \citet{bricken2023towards}, we (i) tie the weights of the pre-encoder bias and the decoder bias, $b_{\mathrm{pre}} = b_{\dec}$; (ii) initialize $b_{\mathrm{pre}}$ to the geometric median of the training set; (iii) initialize $W_{\dec} = W_{\enc}^T$; and (iv) force all decoder rows to have unit norm. 

To minimize the number of ``dead'' latent neurons in the SAE—neurons which are never selected by the $\topk$ filter—we use an auxiliary loss \citep{gao_scaling_2024, oneill_disentangling_2024}. 
The auxiliary loss term provides nonzero gradients for up to $k_{\text{aux}}$ neurons which were not active in the last several forward passes.
We fit these neurons to the residual reconstruction error left by the $\topk$ neurons. 
That is, let $z_i^{\auxk}$ consist of the activations of the top $k_{\text{aux}}$ dead neurons in $z_i$; then
$$\mathcal{L}_{\text{aux}} = ||W_{\dec} z_i^{\auxk} - (e_i - \hat{e}_i)||^2_2,$$
and the full loss, then, is $\mathcal{L} = \mathcal{L}_{\topk} + w_{\text{aux}}\mathcal{L}_{\auxk}$. 

Hyperparameters for the auxiliary loss and other training hyperparameters are given below (identical for all experiments):

\begin{itemize}
    \item $\bm{k_{\text{aux}}}$: $2\cdot k$; that is, up to $2\cdot k$ dead neurons are used to fit the $\topk$ reconstruction error.
    \item \textbf{Dead neuron threshold steps}: 256. This is the number of consecutive steps for which a neuron was not selected by $\topk$ to be considered ``dead.''
    \item \textbf{Auxilliary loss coefficient $w_{\text{aux}}$}: $1/32$. This parameter gives the weight of the auxiliary reconstruction's loss.
    \item \textbf{Batch size}: 512; \textbf{learning rate}: 5e-4; \textbf{gradient clipping threshold}: 1.0.
    \item \textbf{Epochs}: Up to $200$, with early stopping after 5 epochs of validation loss not decreasing.
\end{itemize}

\subsection{Interpreting neurons with LLMs}
\label{sec:hyperparams_neuronlabeling}

There are several parameters for neuron interpretation which we fix across experiments:
\begin{itemize}
    \item \textbf{Interpretation model}: GPT-4o (version 2024-11-20).
    \item \textbf{Temperature}: 0.7.
    \item \textbf{Number of highly-activating examples}: 10.
    \item \textbf{Number of weakly-activating examples}: 10.
    \item \textbf{Maximum word count per example}: 256 (examples longer than this are truncated).
    \item \textbf{Number of candidate interpretations}: 3 (we choose the highest-fidelity interpretation out of 3 candidates).
    \item \textbf{Number of samples to evaluate fidelity}: 200.
\end{itemize}

In Appendix \ref{sec:autointerp_expts}, we describe experiments which justify these choices: they either improve fidelity or have no effect compared to alternate choices. 
We also provide the full interpretation prompt in Figure \ref{fig:interp_prompt_binary}.

There is one hyperparameter which we adjust per dataset: the \textbf{bin percentiles} from which we sample highly-activating and weakly-activating examples for the neuron interpretation prompt.
By default, highly-activating texts are sampled from the top decile of positive activations $[90, 100]$, and weakly-activating texts are sampled from the bottom decile of positive activations $[0, 10]$.
This works well in general, especially for larger datasets (\yelp, \congress).
For smaller datasets like \wiki, the $[90, 100]$ bin produces interpretations which are too specific (e.g., ``articles about Napoleon'', even if the neuron fires on articles about European leaders more generally).
Note that this requires a judgment call: what is ``too specific'' depends on practitioner needs.
For example, on \wiki, we choose bins such that the resulting interpretations are roughly as granular as the topics in the dataset\footnote{Choosing this parameter requires similar decision-making to choosing the minimum cluster size in \bertopic.}. 
We ultimately choose the following percentile bins for highly-activating texts:

\begin{itemize}
    \item \wiki-5, \wiki-15, \headlines: $[80, 100]$.
    \item \bills: $[50, 100]$.
    \item \yelp, \congress: $[90, 100]$.
\end{itemize}

Choosing the \textbf{number of candidate interpretations} depends on budget. 
In Appendix \ref{sec:autointerp_expts} we find that selecting the interpretation with highest fidelity from a pool of 3 candidates increases the resulting fidelity modestly, yet statistically significantly; we therefore use \textbf{3 candidate interpretations} for all experiments.
However, we also ablate this step—that is, generating a single candidate interpretation and using it directly, without computing fidelity. 
On \congress (we did not test on the other datasets), the resulting hypotheses are of similar quality: 14/20 are significant, with AUC 0.70, compared to 15/20 significant and AUC 0.70 when using 3 candidate interpretations\footnote{To illustrate this point, we tested solely on \congress; we expect similarly modest differences on other datasets.}. 
This variant of our method, \textsc{Sae-NoVal}, costs as little as \bertopic (Table \ref{tab:costs}). 

\section{Supporting Experiments}
\label{sec:autointerp_expts}

Our final neuron interpretation procedure for our main experiments is described in \S\ref{sec:methods_main}, with hyperparameters provided in Appendix \ref{sec:hyperparams_neuronlabeling}. 
Here, we describe several experiments to justify our procedure. 
The section is organized as follows.
In \ref{sec:fidelity_predictiveness}, we show that our metric of interpretation fidelity—the F1 score comparing concept annotations against neuron activations—is a useful selection metric to improve downstream hypothesis quality.
In \ref{sec:interp_fidelity}, we test several different interpretation strategies and hyperparameters and measure their impacts on fidelity.
In \ref{sec:performance_losses}, we analyze how much prediction signal is lost at each stage of our method, shedding light on how future work can improve hypothesis quality.

\subsection{Higher-fidelity interpretations improve hypotheses.}
\label{sec:fidelity_predictiveness}

To test whether fidelity is a useful metric to select from different neuron interpretations, we conduct the following experiment on the Yelp dataset:
\begin{enumerate}
    \item Train an SAE (Step 1), and identify the top-20 predictive neurons via Lasso (Step 2).
    \item Generate 10 candidate interpretations (at temperature 0.7, and with different random samples in the interpretation prompt) for each neuron using the hyperparameters in \ref{sec:hyperparams_neuronlabeling} (Step 3).
    \item Compute the \textit{fidelity} for each candidate interpretation: that is, binarizing 100 examples in the top decile of the neuron's positive activations as positives, and 100 examples in the bottom decile of the neuron's positive activations as negatives, compute the F1 score between the GPT-4o-mini annotations according to the interpretation against the binarized neuron activations. 
    \item Store two sets of neuron interpretations: one set includes the interpretations with the best F1 score for each neuron (high-fidelity), and the other set includes the interpretations with the worst F1 score for each neuron (low-fidelity).
    \item Compute annotations on the full heldout set (10K examples) for the high-fidelity hypotheses and the low-fidelity hypotheses, and compare how predictive they are as measured by $R^2$.
\end{enumerate}

This procedure yields three interesting results. 
First, there is enough variation in the 10 candidate interpretations such that the best and worst F1 scores are substantially different: the median best F1 is 0.79, while the median worst F1 is 0.53.
Second, we find that the high-fidelity set is indeed more predictive: in a multivariate OLS, the high-fidelity set achieves an $R^2$ of 0.76, compared to 0.61 for the low-fidelity set.

\begin{figure*}[!h]
\begin{center}
    \includegraphics[width=0.5\linewidth]{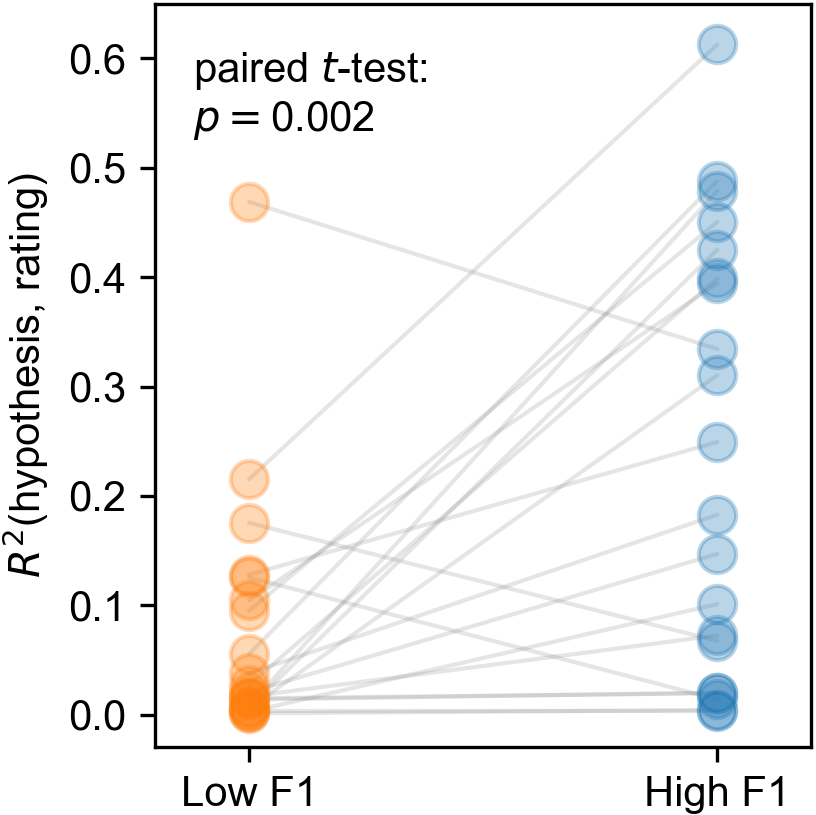}
\end{center}
    \caption{For 17 of the top 20 neurons on \yelp, a higher-fidelity neuron interpretation results in a more predictive hypothesis. 
    Fidelity is measured by the F1 score between the concept annotations and the neuron activations on a sample of 100 highly-activating and 100 weakly-activating examples.
    We generate 10 candidate interpretations for each neuron, and compare the interpretations with best- and worst-F1.
    Predictiveness is measured by computing the $R^2$ between the concept annotations and $Y$ (restaurant rating) for 10K heldout examples.
    }
    \label{fig:fidelity_experiment}
\end{figure*}

Third, most importantly, using a neuron's high-fidelity interpretation results in a more predictive hypothesis, on average.
Figure \ref{fig:fidelity_experiment} shows that, for 17 out of 20 neurons, moving from the low-fidelity interpretation to the high-fidelity interpretation increases the $R^2$ of the resulting hypothesis ($p$-value, paired $t$-test: 0.002).
For example, for one neuron, the high-fidelity interpretation is ``\textit{mentions disputes or issues related to billing or refunds}'' (F1: 0.79), while the low-fidelity interpretation is more generically ``\textit{mentions an attempt by the restaurant or management to address or respond to a problem or complaint}.'' 
The former hypothesis achieves heldout $R^2 = 0.10$ with restaurant rating, while the latter is not significantly correlated with rating ($R^2 = 0.003$).
These experiments show that \textbf{fidelity—as measured via F1 score—is a useful heuristic to choose between multiple candidate neuron interpretations}, when our goal is to produce predictive hypotheses.

Finally, following prior work \citep{bills2023language}, we also tested another metric of fidelity: the $R^2$ between the concept annotations and the neuron activations across 100 top-activating samples and 100 random samples. 
Choosing the top candidate interpretation based on this alternative fidelity metric did not produce hypotheses that were significantly more or less predictive than using our F1 fidelity metric. 

\subsection{Effects of the neuron interpretation procedure on fidelity.} 
\label{sec:interp_fidelity}

Having established that improving interpretation fidelity improves downstream predictiveness, we search across different hyperparameters for neuron interpretation to maximize fidelity. 
Our final procedure is to generate 3 candidate interpretations for each neuron by prompting GPT-4o with 10 examples in the top decile of positive activations vs. 10 examples in the bottom decile of positive activations. 
We use the highest fidelity interpretation out of these 3.
There are two sources of non-determinism: each candidate interpretation is generated with a different random seed to sample examples, and we use an LLM temperature of 0.7.

To compare this strategy against other strategies—or to compare strategy A and strategy B more generally—we compute interpretations, and their F1 scores, for the top 100 neurons on \yelp under both strategies separately.
We run a $t$-test comparing the pairs of resulting F1 scores to test whether one strategy produces consistently higher F1 scores than the other.

We observe the following:
\begin{itemize}
    \item \textbf{Number of candidate interpretations:} Relative to baseline (generating 3 candidate interpretations and taking the one with highest F1 score), generating only a single candidate interpretation is significantly worse (-0.04). 
    Generating 5 candidate interpretations improves significantly on 3 (+0.01 F1, paired $t$-test $p = 0.003$), but the improvement is modest and results in a slower, more expensive method. 
    Therefore, we use only 3 candidate interpretations.
    
    \item \textbf{Top-random vs. high-weak sampling strategy:} Prior work (e.g.~\citet{templeton2024scaling, bills2023language}) interprets SAE neurons by prompting with the top-activating texts and random texts. 
    Relative to our baseline (prompting with highly-activating texts, but not necessarily top-activating; and weakly-activating texts instead of random), this alternate approach has no significant effect on F1 score, though the interpretations tend to be more specific: that is, they have higher precision and lower recall. 
    This is because a neuron's absolute top-activating examples generally share a more specific characteristic than examples sampled randomly from the top decile.
    Though neither strategy wins clearly in terms of F1, we choose high-weak sampling for our experiments because, empirically, top-random produces hypotheses which can be \textit{too} specific (thereby lowering predictiveness); this is discussed further in \ref{sec:hyperparams_neuronlabeling}.
    \textbf{Our code defaults to top-random} for simplicity, but practitioners should treat sampling strategy as a tunable hyperparameter that depends on desired hypothesis granularity.
    
    \item \textbf{Continuous vs. binary prompt structure:} Prior work (e.g.~\citet{zhong_explaining_2024}) prompts the interpreter LLM with a sorted list of texts and their continuous activations. Relative to our baseline (showing separate lists of positive samples and negative samples), this significantly lowers F1 score (-0.06), so we use the binary prompt structure.
    
    \item \textbf{Chain-of-thought:} Prior work (e.g.~\citet{oneill_disentangling_2024}) uses a chain-of-thought reasoning prompt to interpret neurons, where the LLM is encouraged to iteratively discover a concept which all of the positive samples contain and none of the negative samples contain. 
    Relative to baseline, we found that CoT has an insignificant but weakly negative effect (-0.03 mean F1), and also sometimes fails to output any interpretation. 
    However, this is still a promising line for future work: it's possible that while GPT-4o does not reason well, better models or reasoning-specific models (OpenAI o1, DeepSeek R1, \textit{etc.}) will produce higher fidelity labels with chain-of-thought.
    
    \item \textbf{Sample count:} Relative to baseline (prompting with 20 total samples, 10 highly-activating and 10 weakly-activating), prompting with 10 total samples lowers average F1 score by 0.03. Prompting with 50 samples has no effect, so we use 20 to favor shorter prompts.
    
    \item \textbf{Temperature:} Relative to baseline (temperature 0.7), using a lower (0.0) or higher (1.0) temperature has an insignificant but weakly negative effect (-0.01 in mean F1), so we use 0.7.
\end{itemize}

\subsection{Localizing losses in predictive performance to different steps of the method.}
\label{sec:performance_losses}

\ourmethod starts from embeddings and ultimately produces hypotheses, after three steps. 
It is worth asking—how does predictive performance change at each step of our method?
Answering this question points to possible improvements: if the SAE's full hidden representation is much less predictive of restaurant ratings than the input text embeddings, perhaps we need a larger (more expressive) SAE, or more data to train the SAE's representations; meanwhile, if the largest performance loss comes from translating neurons into natural language hypotheses, perhaps we need a stronger LLM for interpretation.

\begin{figure*}[!h]
\begin{center}
    \includegraphics[width=0.6\linewidth]{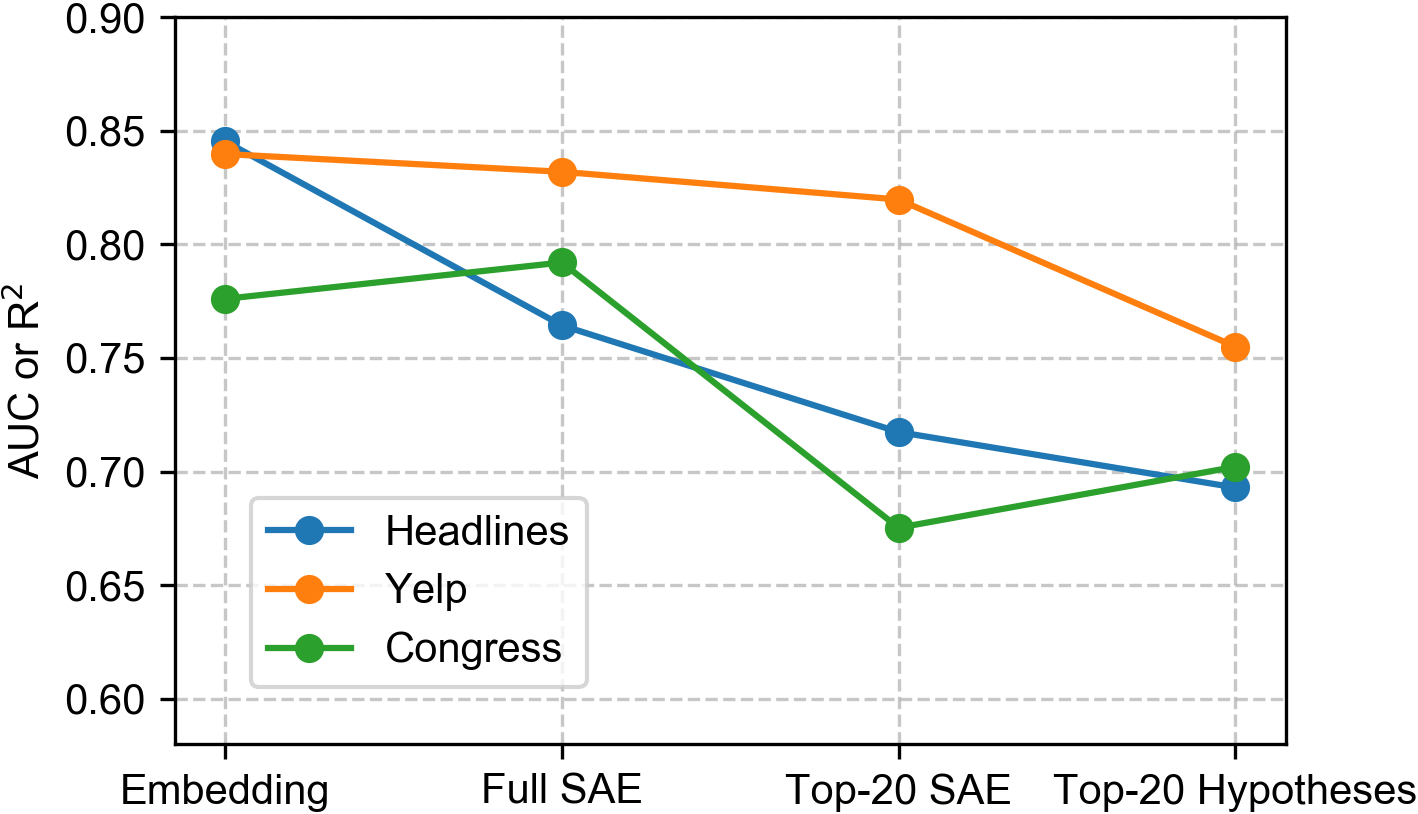}
\end{center}
    \caption{Performance losses at each step of the \ourmethod pipeline. 
    The leftmost point shows the performance of training a supervised single-layer model to predict the outcome using text embeddings; the second point shows the supervised performance using the full SAE representation; the third point shows the supervised performance using the top-20 SAE neurons; the rightmost point shows the supervised performance using the top-20 hypothesis annotations after neuron interpretation. 
    }
    \label{fig:performance_losses}
\end{figure*}

Figure \ref{fig:performance_losses} displays this experiment. 
At each stage, we fit a logistic or linear regression on the training set to predict the labels on \headlines, \yelp, and \congress.
The four stages are: (i) text embeddings (length-1536 vectors of floats); (ii) the full SAE activation matrix, $Z_{\text{SAE}}$ (length-288, 1024, and 4096 vectors of floats for the three tasks, respectively); (iii) the activations of the top-20 selected SAE neurons (length-20 vectors of floats); and (iv) the annotations from the interpreted hypotheses (length-20 vectors of binary 0/1 values).

The trends depend on the dataset. 
On \headlines, we find that the performance drops most at the first stage (-0.08 AUC): the full SAE is significantly less expressive than the original embeddings. This could suggest using a larger SAE, but empirically we find that increasing the SAE size does not improve AUC, perhaps because there is not enough data ($\sim$14K headlines) to make use of more parameters.
On \yelp, we find that performance is mostly retained until neuron interpration, at which point $R^2$ drops by 0.06. 
This suggests some combination of: our neuron interpretations are imperfect, GPT-4o-mini is not annotating the hypotheses faithfully, or binary annotations are not expressive enough to capture the neuron activations (which are continuous).
Finally, on \congress, the only drop comes from using the top-20 neurons to predict the label instead of the full SAE representation (4096 neurons). 
This implies that 20 features aren't expressive enough; or, more optimistically, that there are more than 20 distinct hypotheses to discover on this dataset (see \S\ref{sec:novelty}). 
Remarkably, there are no losses in the interpretation step on the \congress dataset, suggesting that on this dataset our interpretations are very high-fidelity.
Ultimately, this analysis can be a useful diagnostic tool to direct one's efforts when trying to improve hypothesis quality: losses from embedding $\Rightarrow$ full SAE suggest that the dataset and/or the SAE are too small; losses from full SAE $\Rightarrow$ top-$H$ SAE suggest that $H$ is too small; losses from top-$H$ SAE $\Rightarrow$ top-$H$ hypotheses suggest that the interpretation step can be improved.
Meanwhile, if the top-$H$ hypotheses achieve similar AUC to the underlying embeddings, this implies that there may be no more ``juice to squeeze'' in the dataset.

\section{Hyperparameters for Baseline Methods}

\paragraph{\bertopic \citep{grootendorst_bertopic_2022}.}
To run \bertopic\footnote{\href{https://github.com/MaartenGr/BERTopic}{https://github.com/MaartenGr/BERTopic}}, we feed in the same OpenAI  embeddings that we use for \ourmethod, and otherwise use default choices for the algorithm: UMAP for dimensionality reduction, HDBSCAN for clustering, and c-TF-IDF to compute the top words associated with each topic.
We tune the minimum cluster size hyperparameter in \bertopic, which controls topic granularity. 
The default value is 10, which often produces topics which are too granular (e.g., relative to the underlying granularity of the subtopics on \wiki and \bills). 
We test all values in $\{\text{10, 20, 50, 100, 200, 500}\}$. 
We choose this parameter by maximizing the validation performance of the $L_1$-regularized predictor which uses $H$ features; this is identical to how we choose hyperparameters for \ourmethod.
We use default values of all other hyperparameters, which is standard in prior work (e.g., \citet{pham_topicgpt_2024}). 
After running \bertopic, we label topics using the prompt Figure \ref{fig:bertopic_prompt}.

\paragraph{\nlparam \citep{zhong_explaining_2024}.} 
We run \nlparam using the publicly available implementation\footnote{\href{https://github.com/ruiqi-zhong/nlparam}{https://github.com/ruiqi-zhong/nlparam}}. 
We set the number of candidate labels per bottleneck layer neuron to 3 (the same as our method), and we use the same text embeddings as our method (OpenAI's text-3-embedding-small).
\citet{zhong_explaining_2024} release two different prompts, one for ``simple'' hypotheses and one for ``complex'' hypotheses; following their work, we use the ``simple'' prompt for our 3 synthetic tasks and the ``complex'' prompt for our 3 real-world tasks.
For all other hyperparameters, we use default values.

\paragraph{\hypogenic \citep{zhou_hypothesis_2024}.} We run \hypogenic using the publicly available implementation\footnote{\href{https://github.com/ChicagoHAI/hypothesis-generation}{https://github.com/ChicagoHAI/hypothesis-generation}}. 
We use the default parameters as defined in examples/generation.py. Following the format in their example tasks, we write custom prompts in a consistent format describing each our of tasks. 
The method uses the same LLM for hypothesis generation and hypothesis scoring, and we run all experiments using GPT-4o-mini (the default model in their code repository, and also the model that we use for scoring hypotheses in \ourmethod).

\section{Qualitative Evaluation}\label{sec:qualeval}

To confirm that our hypotheses pass a ``sniff test,'' we recruit three computational social science researchers to perform a qualitative evaluation. 
These researchers are not authors on our study.
We collect ratings for two attributes, following \citet{lam2024conceptInduction}: (1) \textit{helpfulness} and (2) \textit{interpretability}. 
We explain these to the raters as follows:
\begin{enumerate}
    \item Helpful: Does this hypothesis help you understand what predicts the target variable? If you were studying this dataset, is this a hypothesis you would want to and could explore further? Rate 1 if yes, 0 if no or only a little.
    \item Interpretable: When you read the hypothesis, is it clear what it means? Can you easily and unambiguously apply it to new examples? Rate 1 if yes, 0 if no or it would often be subjective.
\end{enumerate}

We collect 3 annotations for each of the significant hypotheses outputted by any method on the \headlines and \congress datasets. (We removed the \yelp dataset to reduce annotator load.)
Hypotheses are shown to annotators in random order, i.e., they were not grouped by method.
We aggregate labels by taking the majority label. 

\section{Proofs}\label{sec:proof}

\begin{proof}[Proof of \Cref{prop:errors-in-variables}]
Set $p_{ab}:=\Pr[\hat{Z}=a, Z=b]$ and $y_{ab}:=\EE[Y|\hat{Z}=a, Z=b]$
for $a,b\in \{0,1\}.$ Then
\begin{align}
    \EE[Y|\hat{Z}=1] &= \frac{y_{11}p_{11} + y_{10}p_{10}}{p_{11}+p_{10}}\\
    \EE[Y|Z=1] &= \frac{y_{11}p_{11} + y_{01}p_{01}}{p_{11}+p_{01}}.
\end{align}
Therefore,
\begin{align}
    \EE[Y|\hat{Z}=1] - \EE[Y|Z=1] &= \frac{y_{11}p_{11} + y_{10}p_{10}}{p_{11}+p_{10}} - \frac{y_{11}p_{11} + y_{01}p_{01}}{p_{11}+p_{01}}\\
    &= y_{11}\left(\frac{p_{11}}{p_{11}+p_{10}} - \frac{p_{11}}{p_{11}+p_{01}}\right) + \frac{y_{10}p_{10}}{p_{11}+p_{10}} - \frac{y_{01}p_{01}}{p_{11}+p_{01}}\\
    &= y_{11}\left(\frac{p_{01}}{p_{11}+p_{01}} - \frac{p_{10}}{p_{11}+p_{10}}\right) + \frac{y_{10}p_{10}}{p_{11}+p_{10}} - \frac{y_{01}p_{01}}{p_{11}+p_{01}}\label{eq:cat}
\end{align}
Since $y_{10}\le 1$ and $y_{01}\ge 0,$ \eqref{eq:cat} is bounded from above by
\begin{align}
    y_{11}\left(\frac{p_{01}}{p_{11}+p_{01}} - \frac{p_{10}}{p_{11}+p_{10}}\right) + \frac{p_{10}}{p_{11} + p_{10}}.\label{eq:dog}
\end{align}
If $\frac{p_{01}}{p_{11}+p_{01}} - \frac{p_{10}}{p_{11}+p_{10}} \ge 0$, \eqref{eq:dog} is bounded from above by
\begin{align}
    &\left(\frac{p_{01}}{p_{11}+p_{01}} - \frac{p_{10}}{p_{11}+p_{10}}\right) + \frac{p_{10}}{p_{11} + p_{10}}
    = \frac{p_{01}}{p_{11}+p_{01}},
\end{align}
since $y_{11} \le 1.$ Otherwise, \eqref{eq:dog}  is bounded from above by
\begin{align}
    \frac{p_{10}}{p_{11}+p_{10}}.
\end{align}
Also note that \eqref{eq:cat} is bounded from below by
\begin{align}
    y_{11}\left(\frac{p_{11}}{p_{11}+p_{10}} - \frac{p_{11}}{p_{11}+p_{01}}\right) - \frac{p_{01}}{p_{11} + p_{10}} = y_{11}\left(\frac{p_{01}}{p_{11}+p_{01}} - \frac{p_{10}}{p_{11}+p_{10}}\right) - \frac{p_{01}}{p_{11} + p_{10}}.\label{eq:dog-2}
\end{align}
If $\frac{p_{01}}{p_{11}+p_{01}} - \frac{p_{10}}{p_{11}+p_{10}} \le 0$, \eqref{eq:dog-2} is bounded from below by
\begin{align}
    &\left(\frac{p_{01}}{p_{11}+p_{01}} - \frac{p_{10}}{p_{11}+p_{10}}\right) - \frac{p_{01}}{p_{11} + p_{10}}
    = -\frac{p_{10}}{p_{11}+p_{10}},
\end{align}
since $y_{11} \le 1.$ Otherwise, \eqref{eq:dog-2}  is bounded from below by
\begin{align}
    -\frac{p_{01}}{p_{11}+p_{01}}.
\end{align}
Therefore, 
\begin{equation}
    |\EE[Y|\hat{Z}=1] - \EE[Y|Z=1]|\le \max\left\{\frac{p_{10}}{p_{11} + p_{10}}, \frac{p_{01}}{p_{11} + p_{01}}\right\}.
\end{equation}

Analogously, we can bound $|\EE[Y|\hat{Z}=0] - \EE[Y|Z=0]|$ by 
\begin{equation}
    \max\left\{\frac{p_{10}}{p_{00} + p_{10}}, \frac{p_{01}}{p_{00} + p_{01}}\right\}.
\end{equation}
Notice that
\begin{align}
    \frac{p_{10}}{p_{00} + p_{10}} &= \frac{p_{10}}{p_{11}+p_{10}}\cdot \frac{p_{11}+p_{10}}{p_{00} + p_{10}}\\
    &=\frac{p_{10}}{p_{11}+p_{10}}\cdot \frac{\Pr[\hat{Z}=1]}{\Pr[Z=0]}
\end{align}
Similarly,
\begin{align}
    \frac{p_{01}}{p_{00} + p_{01}} &= \frac{p_{01}}{p_{11}+p_{01}}\cdot \frac{p_{11}+p_{01}}{p_{00} + p_{01}}\\
    &=\frac{p_{01}}{p_{11}+p_{01}}\cdot \frac{\Pr[Z=1]}{\Pr[\hat{Z}=0]}
\end{align}
Set $p:=\min\{\Pr[\hat{Z}=0], \Pr[Z=0]\}$. Then $\Pr[\hat{Z}=0], \Pr[Z=0]\ge p$ and $\Pr[\hat{Z}=1], \Pr[Z=1]\le 1-p$
\begin{equation}
    \frac{\Pr[\hat{Z}=1]}{\Pr[Z=0]}, \frac{\Pr[Z=1]}{\Pr[\hat{Z}=0]} \le \frac{1-p}{p}. 
\end{equation}
It follows that
\begin{equation}
    \left|S(\hat{Z}) - S(Z)\right| = \left|(\EE[Y|\hat{Z}=1] - \EE[Y|\hat{Z}=0]) - (\EE[Y|Z=1] - \EE[Y|Z=0])\right|
\end{equation}
is at most
\begin{align}
    &\left(1 + \frac{1-p}{p}\right)\max\left\{\frac{p_{10}}{p_{11}+p_{10}}, \frac{p_{01}}{p_{11}+p_{01}}\right\}
    = \frac{1}{p}\max\left\{\frac{p_{10}}{p_{11}+p_{10}}, \frac{p_{01}}{p_{11}+p_{01}}\right\}.
\end{align}
The result follows by noting that $p:=\min\{\Pr[\hat{Z}=0], \Pr[Z=0]\}$ and $\frac{p_{10}}{p_{11}+p_{10}}, \frac{p_{01}}{p_{11}+p_{01}} = 1 - \text{ recall}, 1 - \text{ precision}.$
\end{proof}

\section{Prompts}

\begin{figure*}[!htb]
\centering\small
\begin{tcolorbox}[colback=blue!5!white, colframe=blue!75!black, boxrule=0.5mm, arc=4mm, boxsep=5pt, width=\textwidth]
You are a machine learning researcher who has trained a neural network on a text dataset. You are trying to understand what text features cause a specific neuron in the neural network to fire. \\

You are given two sets of SAMPLES: POSITIVE SAMPLES that strongly activate the neuron, and NEGATIVE SAMPLES from the same distribution that do not activate the neuron. Your goal is to identify a feature that is present in the positive samples but absent in the negative samples. \\
Example features could be: \\
- ``uses multiple adjectives to describe colors'' \\
- ``describes a patient experiencing seizures or epilepsy'' \\
- ``contains multiple single-digit numbers'' \\

POSITIVE SAMPLES: \\
---------------- \\
\{positive\_samples\} \\
---------------- \\

NEGATIVE SAMPLES: \\
---------------- \\
\{negative\_samples\} \\
---------------- \\

Rules about the feature you identify: \\
- The feature should be objective, focusing on concrete attributes rather than abstract concepts. \\
- The feature should be present in the positive samples and absent in the negative samples. Do not output a generic feature which also appears in negative samples. \\
- The feature should be as specific as possible, while still applying to all of the positive samples. For example, if all of the positive samples mention Golden or Labrador retrievers, then the feature should be ``mentions retriever dogs'', not ``mentions dogs'' or ``mentions Golden retrievers''. \\

Do not output anything else. Your response should be formatted exactly as shown in the examples above. Please suggest exactly one description, starting with ``-'' and surrounded by quotes ``''. Your response is: \\
- ``
\end{tcolorbox}
\caption{The prompt we use to interpret SAE neurons. 
The positive samples are texts that strongly activate the neuron, while the negative samples are texts that weakly activate the neuron. 
The ``example features'' section can be edited to produce features in a different format.
}
\label{fig:interp_prompt_binary}
\end{figure*}

\begin{figure*}[!htb]
\centering\small
\begin{tcolorbox}[colback=blue!5!white, colframe=blue!75!black, boxrule=0.5mm, arc=4mm, boxsep=5pt, width=\textwidth]
Is text\_a and text\_b similar in meaning? Respond with yes, related, or no. \\
\\
Here are a few examples. \\
\\
Example 1: \\
text\_a: has a topic of protecting the environment \\
text\_b: has a topic of environmental protection and sustainability \\
output: yes \\
\\
Example 2: \\
text\_a: has a language of German \\
text\_b: has a language of Deutsch \\
output: yes \\
\\
Example 3: \\
text\_a: has a topic of the relation between political figures \\
text\_b: has a topic of international diplomacy \\
output: related \\
\\
Example 4: \\
text\_a: has a topic of the sports \\
text\_b: has a topic of sports team recruiting new members \\
output: related \\
\\
Example 5: \\
text\_a: has a named language of Korean \\
text\_b: uses archaic and poetic diction \\
output: no \\
\\
Example 6: \\
text\_a: has a named language of Korean \\
text\_b: has a named language of Japanese \\
output: no \\
\\
Example 7: \\
text\_a: describes an important 20th century historical event \\
text\_b: describes a 20th century European politician \\
output: no \\
\\
Target: \\
text\_a: \{text\_a\} \\
text\_b: \{text\_b\} \\
output:
\end{tcolorbox}
\caption{The prompt we use to evaluate surface similarity between the inferred hypotheses and reference topics. The prompt is lightly edited from \citet{zhong_explaining_2024}.
The few-shot examples demonstrate three levels of similarity: equivalent meaning (yes), related concepts (related), and unrelated concepts (no).}
\label{fig:surface_similarity_prompt}
\end{figure*}

\begin{figure*}[!htb]
\centering\small
\begin{tcolorbox}[colback=blue!5!white, colframe=blue!75!black, boxrule=0.5mm, arc=4mm, boxsep=5pt, width=\textwidth]
Check whether the TEXT satisfies a PROPERTY. Respond with Yes or No with an explanation that discusses the evidence from the TEXT (at most a sentence). When uncertain, output No. \\
\\
Example 1: \\
PROPERTY: ``mentions a natural scene.'' \\
TEXT: ``I love the way the sun sets in the evening.'' \\
Output: Yes. ``Sun sets'' are clearly natural scenes. \\
\\
Example 2: \\
PROPERTY: ``writes in a 1st person perspective.'' \\
TEXT: ``Jacob is smart.'' \\
Output: No. This text is written in a 3rd person perspective. \\
\\
Example 3: \\
PROPERTY: ``is better than group B.'' \\
TEXT: ``I also need to buy a chair.'' \\
Output: No. It is unclear what the PROPERTY means (e.g., what does group B mean?) and doesn't seem related to the text. \\
\\
Example 4: \\
PROPERTY: ``mentions that the breakfast is good on the airline.'' \\
TEXT: ``The airline staff was really nice! Enjoyable flight.'' \\
Output: No. Although the text appreciates the flight experience, it DOES NOT mention about the breakfast. \\
\\
Example 5: \\
PROPERTY: ``appreciates the writing style of the author.'' \\
TEXT: ``The paper absolutely sucks because its underlying logic is wrong. However, the presentation of the paper is clear and the use of language is really impressive.'' \\
Output: Yes. Although the text dislikes the paper, it says ``the presentation of the paper is clear'', so it DOES like the writing style. \\
\\
Example 6: \\
PROPERTY: ``has a formal style; specifically, the language in the text is relatively formal, complex and academic. For example, 'represent whom and which''' \\
TEXT: ``investigates formation of nominalization'' \\
Output: Yes. ``formation'' and ``nominalization'' are abstract and complex nouns. \\
\\
Example 7: \\
PROPERTY: ``refers to historical dates; specifically, there are references to years or specific dates in the text. For example, 'Obama was born on August 4, 1961.''' \\
TEXT: ``A member of the Democratic Party, he was the first African-American president of the United States.'' \\
Output: No. The text does not mention date. \\
\\
Now complete the following example - Respond with Yes or No with an explanation that discusses the evidence from the TEXT. When uncertain, output No. \\
\\
PROPERTY: \{hypothesis\} \\
TEXT: \{text\} \\
Output:
\end{tcolorbox}
\caption{The prompt we use to compute hypothesis annotations with GPT-4o-mini to evaluate hypothesis quality. The prompt is lightly edited from \citet{zhong_explaining_2024}.}
\label{fig:concept_annotation_prompt}
\end{figure*}

\begin{figure*}[!htb]
\centering\small
\begin{tcolorbox}[colback=blue!5!white, colframe=blue!75!black, boxrule=0.5mm, arc=4mm, boxsep=5pt, width=\textwidth]
You are a machine learning researcher analyzing clusters from a topic model. You need to identify the key distinguishing features of documents in a specific topic cluster. \\
Example features could be: \\
- ``uses multiple adjectives to describe colors'' \\
- ``describes a patient experiencing seizures or epilepsy'' \\
- ``contains multiple single-digit numbers'' \\
\\
These are the top keywords describing the topic: \{topic\_keywords\}. \\
\\
You will also see a sample of documents belonging to this topic and a random sample of documents from other topics from the same dataset. \\
\\
TOPIC DOCUMENTS (a sample of documents belonging to this topic): \\
---------------- \\
\{topic\_documents\} \\
---------------- \\
\\
RANDOM DOCUMENTS (a sample of documents from other topics): \\
---------------- \\
\{random\_documents\} \\
---------------- \\
\\
Your goal is to create a short, specific label that captures what makes the topic documents different from the contrasting documents. \\
\\
The label should be: \\
1. As specific as possible, while still applying to all topic documents \\
2. Focused on concrete features rather than abstract concepts \\
3. Based on characteristics present in the topic documents but not in the contrasting documents \\
\\
Do not output anything else. Your response should be a single topic label starting with ``-'' and surrounded by quotes ``''. Your response is: \\
- ``
\end{tcolorbox}
\caption{The prompt we use to generate interpretable topic labels for \bertopic. 
The LLM is prompted with the keywords that distinguish the topic from others, and also with two sets of documents: 10 documents belonging to the target topic, and a random sample of 10 documents from other topics.
This labeling approach is inspired by \citet{grootendorst_llm_2024}, but we designed the prompt ourselves (intending it to be similar, where possible, to our neuron interpretation prompt).
}
\label{fig:bertopic_prompt}
\end{figure*}

\end{document}